\pdfoutput=1
\documentclass[11pt]{article}

\usepackage[]{acl}

\usepackage{times}
\usepackage{latexsym}

\usepackage[T1]{fontenc}
\usepackage[utf8]{inputenc}
\usepackage{microtype}
\usepackage{inconsolata}

\usepackage{adjustbox}
\usepackage{booktabs}
\usepackage{makecell}
\usepackage{multirow}
\usepackage{amsmath}
\usepackage{comment}
\usepackage{caption}
\usepackage{subcaption}
\usepackage{cleveref}
\usepackage{pifont}
\usepackage{xcolor} 
\usepackage{hyperref}
\usepackage{listings}
\usepackage{amssymb}

\definecolor{mygreen}{RGB}{0, 158, 115} 
\newcommand{\cmark}{\textcolor{mygreen}{\ding{51}}}  
\definecolor{myred}{RGB}{204, 121, 167} 
\newcommand{\xmark}{\textcolor{myred}{\ding{55}}}
\definecolor{myorange}{RGB}{230, 159, 0} 

\def\eg{e.g.,~}

\def\ie{i.e.,~}

\newcommand{\dataset}{ProMQA}

\def \submission {}
\ifx \submission \undefined

\else

\fi

\title{
    \dataset{}: Question Answering Dataset for\\Multimodal Procedural Activity Understanding
}

\author{
     Kimihiro Hasegawa$^{1}$ \hspace{0.4cm} Wiradee Imrattanatrai$^{2}$ \hspace{0.4cm} Zhi-Qi Cheng$^{1}$ \hspace{0.4cm} Masaki Asada$^{2}$ \\
    {\bf Susan Holm$^{1}$} \hspace{0.4cm} {\bf Yuran Wang$^{1}$} \hspace{0.4cm} {\bf Ken Fukuda$^{2}$} \hspace{0.4cm} {\bf Teruko Mitamura$^{1}$} \\
    $^{1}$Language Technologies Institute, Carnegie Mellon University \\
    $^{2}$National Institute of Advanced Industrial Science and Technology (AIST) \\
    \texttt{kimihiro@cs.cmu.edu}
}

\begin{document}
\maketitle
\begin{abstract}
Multimodal systems have great potential to assist humans in procedural activities, where people follow instructions to achieve their goals. 
Despite diverse application scenarios, systems are typically evaluated on traditional classification tasks, e.g., action recognition or temporal action segmentation.
In this paper, we present a novel evaluation dataset, \dataset{}, to measure system advancements in application-oriented scenarios.
\dataset{} consists of 401 multimodal procedural QA pairs on user recording of procedural activities, i.e., cooking, coupled with their corresponding instructions/recipes.
For QA annotation, we take a cost-effective human-LLM collaborative approach, where the existing annotation is augmented with LLM-generated QA pairs that are later verified by humans.
We then provide the benchmark results to set the baseline performance on \dataset{}.
Our experiment reveals a significant gap between human performance and that of current systems, including competitive proprietary multimodal models. 
We hope our dataset sheds light on new aspects of models' multimodal understanding capabilities.\footnote{Code and data are available at \url{https://github.com/kimihiroh/promqa}.}
\end{abstract}


\section{Introduction}
\label{sec:intro}

\begin{figure*}[t]
    \centering
    \includegraphics[width=\textwidth]{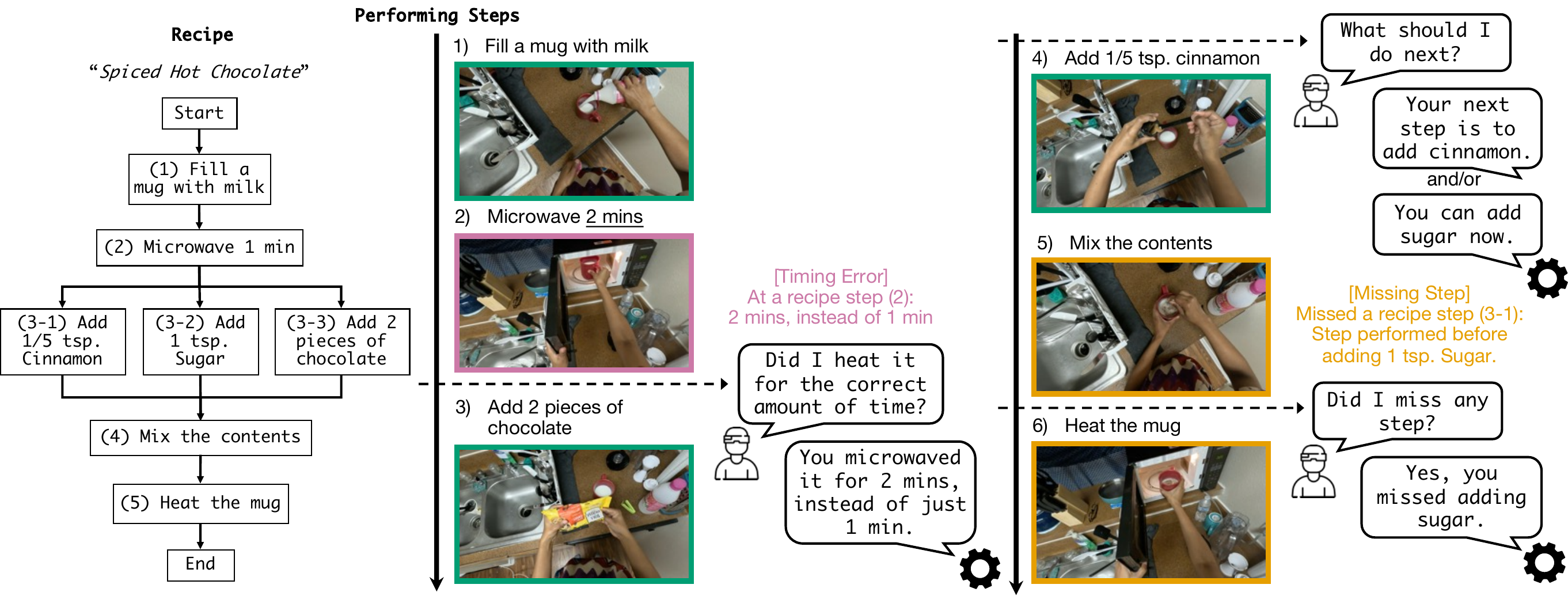}
    \caption{Illustration of a system supporting a user in a procedural activity. The left graph is the recipe and the columns of images are screenshots of the user's actions in chronological order. During the activity, the user makes two mistakes. One is \textit{a timing error}, where the user sets a longer time than required for microwaving (\textcolor{myred}{red}). The other is \textit{a missing step}, where the user skips adding sugar (\textcolor{myorange}{yellow} borders for steps after the missing step). Steps with \textcolor{mygreen}{green} borders do not have any errors. QAs are occurring at each divider's position.}
    \label{fig:overview}
\end{figure*}

Procedures are human knowledge of experience that enables one to obtain an expected outcome without much trial and error. 
Yet, following procedures (\ie a set of instructions), itself requires skills such as, in cooking~\citep{peddi2023captaincook4d}, assembly~\citep{sener2022assembly101}, or surgery~\citep{laura2016virtual}, among others.
In supporting such user activities, current evolving multimodal foundation models like GPT-4o~\citep{openai2024gpt4o} and Claude 3.5 Sonnet~\citep{anthropic2024introducing} have great potential by monitoring the situation through the perception of a user's wearable device.
Despite such diverse application scenarios, existing studies typically provide traditional, but less practical evaluation testbeds.
To support an application-oriented evaluation, we present a novel multimodal question-answering (QA) dataset for understanding procedural activity, produced by our cost-effective human-LLM collaborative approach.

When supporting procedural activities, an assistant should comprehend information from multiple sources:
1) Actual process from their perception;
2) Each step and the overall flow from instructions.
For instance, in cooking, answering ``\textit{What is the next step now?}'' requires an assistant to recognize which steps have been completed until ``\textit{now}'' from its video recording and identify what else/next should be done from its recipe.
Assuming recipes are typically written in text, assistants receive multimodal information of \textit{how one did it} as video and \textit{how one should do it} as text.
Prior work has explored the task in a text-only, unimodal setting, where a user verbalizes all of their actions~\citep{le-etal-2023-improved}.
However, it is not ideal in practice as a beginning cook might give misleading explanations that cannot be corrected by a system without raw information (video) about the actual process.

\autoref{fig:overview} illustrates how one receives cooking support from a system in a reactive manner.
Tailoring toward such a practical scenario, we formulate our task as QA so that multimodal capabilities can be evaluated directly on the downstream task (\S\ref{ssec:task}).
In contrast, prior work traditionally tackles visual action understanding as action recognition and temporal action segmentation~\citep{Kuehne_2014_CVPR, yansong2019coin, Ding2022TemporalAS}.
We argue that these tasks are suboptimal to evaluate procedural activity assistants as they are subtasks of such procedural activity support.

In this work, we present a novel dataset, \textbf{\dataset} (\textbf{Pro}cedural \textbf{M}ultimodal \textbf{Q}uestion \textbf{A}nswering), to evaluate models' capabilities of understanding procedural activities in multimodal settings ({\S\ref{sec:dataset}}).
Our work is motivated by the fact that a well-adapted testbed is indispensable and can stimulate system development.
In the dataset construction, we repurposed videos and recipes from the existing CaptainCook4D~\citep{peddi2023captaincook4d} dataset.
Then, for QA annotation, we employ a human-LLM collaborative approach, where LLMs first generate QA pairs and humans verify them to ensure the quality, inspired by the recent advances in synthetic data generation~\citep{mangalam2023egoschema} (\S\ref{sec:annotation}).
While LLMs cost-effectively generate candidate QA pairs, the manual verification process ensures the quality of the resulting dataset.
Specifically, among 500 generated QA pairs, around 80\% were retained with additional human-written answers through the verification. 
Finally, to establish the baseline performance, we benchmark the following approaches: unimodal models, Socratic models~\citep{zeng2022socraticmodels}, and both open and proprietary multimodal models.
Our benchmark experiments reveal that, while humans can reasonably perform the task, the dataset is challenging even for proprietary multimodal models that show strong performance on other vision-language tasks (\S\ref{sec:benchmark}).

Our contributions are three-fold.
First, we define a novel multimodal QA task and present the dataset, \dataset{}, for procedural activity understanding under a permissive license.\footnote{Apache 2.0}
Second, we propose a human-LLM collaborative approach for cost-efficient QA annotation.
Third, we provide benchmark results to encourage further research on this task.
\section{\dataset}
\label{sec:dataset}

\begin{table*}[!t]
\centering
\fontsize{8}{9}\selectfont
\begin{tabular}{c c c c c c c c}
    \toprule
    \makecell[c]{Dataset\\Name} & Multimodal & Video & Procedural & \makecell[c]{Explicit\\Instruction} & QA & \makecell[c]{Open\\Vocab} & \makecell[c]{LLM\\Scoring}\\
    \midrule
    Assembly101~\citep{sener2022assembly101} & \cmark & \cmark & \cmark & \xmark & \xmark & \xmark & \xmark \\
    IndustReal~\citep{schoonbeek2024industreal} & \cmark & \cmark & \cmark & \xmark & \xmark & \xmark & \xmark \\
    YouCook2~\citep{Zhou_Xu_Corso_2018} & \cmark & \cmark & \cmark & \cmark & \xmark & \xmark & \xmark \\
    CaptainCook4d ~\citep{peddi2023captaincook4d} & \cmark & \cmark & \cmark & \cmark & \xmark & \xmark & \xmark \\
    How2QA~\citep{li-etal-2020-hero} & \cmark & \cmark & \cmark & \xmark & \cmark & \xmark & \xmark \\
    MMBench~\citep{mmbench} & \cmark & \xmark & \xmark & \xmark & \cmark & \cmark & \cmark \\
    EgoSchema~\citep{mangalam2023egoschema} & \cmark & \cmark & \xmark & \xmark & \cmark & \xmark & \xmark \\
    GazeVQA~\citep{ilaslan-etal-2023-gazevqa} & \cmark & \cmark & \cmark & \xmark & \cmark & \xmark & \xmark   \\
    OpenEQA~\citep{majumdar2024openeqa} & \cmark & \cmark & \xmark & \xmark & \cmark & \cmark & \cmark \\
    \midrule
    \dataset{} (Ours) & \cmark & \cmark & \cmark & \cmark & \cmark & \cmark & \cmark \\
    \bottomrule
\end{tabular}
\caption{Our dataset vs. similar multimodal benchmarks}
\label{tab:comparison-dataset}
\end{table*}

\begin{table*}[!t]
\fontsize{8}{9}\selectfont
\begin{tabular}{l p{10cm}}
    \toprule
    Criteria \& Example & Explanation \\
    \midrule 
    \multicolumn{2}{l}{\underline{Multimodal}} \\[0.2em]
    \multirow{1}{*}{{\cmark}~~ What is the next step now?} & This is multimodal because it requires matching the completed steps from the recording to the instructions in order to identify the possible next steps.\\
    \multirow{1}{*}{{\xmark}~~ What am I supposed to do after X?} & This is not multimodal because it can be answered by simply checking the instructions.\\
    \multirow{1}{*}{{\xmark}~~ What did I do after X?} & This is not multimodal because it can be answered by simply checking the recording. \\
    \midrule
    \multicolumn{2}{l}{\underline{Procedural}} \\[0.2em]
    {\cmark}~~ Did I measure X correctly? & This is procedural because it asks specifically about a step. \\
    \multirow{1}{*}{{\xmark}~~ What is the color of the tablespoon?} & This is not procedural because it asks for the static characteristic of a tool.\\
    \midrule
    \multicolumn{2}{l}{\underline{No External Knowledge}} \\[0.2em]
    {\cmark}~~ Did I use the correct tool to measure X? & Suppose the instructions provide sufficient details about the measurement tool, it can be answered using the instructions and the recording, without requiring external knowledge. \\
    {\xmark}~~ Can I replace zucchini with cucumber? & Suppose the recipe does not mention possible replacements, it is unanswerable from the given information. External knowledge would be required to find an answer.\\
    \bottomrule
\end{tabular}
\caption{Criteria of our target multimodal procedural questions with cooking-context examples. Our target questions require both instructions and recordings to answer (multimodal), which are about either the process or each step (procedural) and are answerable from given information (no external knowledge).}
\label{tab:multimodal-procedural-question}
\end{table*}
\begin{table*}[!t]
\fontsize{8}{9}\selectfont
\centering
\begin{tabular}{r | l l }
    \toprule
    \multicolumn{1}{c |}{Question type} & Target & Example question \\
    \midrule 
    \multicolumn{3}{l}{\underline{Process-level}} \\
    Missing & Missing recipe steps & Did I miss any steps so far? \\
    Next & Next recipe steps & What is the next step now? \\
    Order & Errors w.r.t. recipe step ordering & Should I have done any steps in a different order? \\
    \midrule
    \multicolumn{3}{l}{\underline{Step-specific}} \\
    Measurement & \makecell[l]{Errors in measurement (e.g., 2 cups instead of 1 cup)} & Did I measure water correctly? \\
    Preparation & \makecell[l]{Other errors in preparation (e.g., cilantro instead of oregano)} & Did I add the correct spice? \\
    Technique & \makecell[l]{Errors in cooking technique (e.g., chop instead of slice)} & Did I prepare onion correctly? \\
    Temperature & \makecell[l]{Errors in temperature (e.g., high instead of low)} & Was the heat level correct? \\
    Timing & \makecell[l]{Errors in duration (e.g., 2 min instead of 5 min)} & Did I microwave it for long enough? \\
    \bottomrule
\end{tabular}
\caption{Question categories and types with their corresponding target phenomenon and example questions.}
\label{tab:question-type}
\end{table*}

Our goal is to facilitate the development of procedural-activity support systems.
\dataset{} consists of 401 multimodal procedural QA pairs that require both recipes and video recordings to answer.
It is constructed with our human-LLM collaborative approach on top of existing cooking recording and annotation (\S\ref{sec:annotation}).
In~\autoref{tab:comparison-dataset}, we compare our dataset with similar multimodal datasets.
Our dataset uniquely supports the assessment of multimodal procedural activity understanding as the QA task, which can serve as a testbed to advance the model's multimodal procedural activity understanding. 

\subsection{Task Formulation}
\label{ssec:task}
We chose QA as our formulation to better reflect how users seek information and advice in practical situations.
A model takes as input a cooking instruction $recipe$, a recording of a user's activity $video$, and a question $q$, and then, outputs an answer $a$ as natural language. 
A $recipe$ is represented as a directed acyclic graph of recipe steps, whereas a $video$ contains a pile of frames. 
In this work, we treat each QA pair independently, instead of formulating it as dialogue, to focus on reasoning capability, and leave it for future work on how to extend to further practical dialogue settings.
We also note that ``instruction'' and ``recipe'', and ``recording'' and ``video'', are used interchangeably.

\subsection{Multimodal Procedural QA}
\label{ssec:multimodal-procedural-qa}

In \dataset{}, we specifically target multimodal questions about procedural activities. 
Multimodal questions require both instructions and recordings to derive answers, while procedural questions pertain to either individual steps or multiple-step sequences.
In addition, we only retain answerable questions without requiring external or inherent knowledge to emphasize multimodal reasoning capabilities over the provided information. 
\autoref{tab:multimodal-procedural-question} provides examples that distinguish our target from relevant but out-of-scope questions.

Among valid multimodal procedural questions, we categorize them into two groups, where each is further divided into specific question types, following CaptainCook4D.
\textbf{Process-level questions} focus on multiple steps: \textit{missing}, \textit{next}, and \textit{order}. 
\textbf{Step-specific questions} are questions about individual steps: \textit{measurement}, \textit{preparation}, \textit{technique}, \textit{temperature}, and \textit{timing}.
Examples of each type and their descriptions can be found in~\autoref{tab:question-type}.


\begin{table*}[!t]
\centering
\setlength{\tabcolsep}{4pt}
\fontsize{8}{8}\selectfont
\begin{tabular}{cccccccccc}
    \toprule
    \makecell[c]{\#example\\(\#question)} & \makecell[c]{\#distinct \\ recipe} & \makecell[c]{avg. \#steps/\\recipe} & \makecell[c]{\#distinct \\ recording} & \makecell[c]{avg. length \\ of recording} & \makecell[c]{avg. \#steps/\\recording} &  \makecell[c]{avg. \#answers/\\question} & \makecell[c]{avg. \#words/\\question} & \makecell[c]{avg. \#words/\\answer} \\ \midrule
    401 & 24 & 14.3 & 231 & 6m47s & 6.4 & 1.9 & 8.9 & 11.8 \\
    \bottomrule
\end{tabular}
\caption{Statistics of \dataset}
\label{tab:general-stats}
\end{table*}

Answers are categorized into three groups.
Suppose a user asked a question, \eg ``\textit{What should I do next?}''.
\textbf{Direct answers} directly address the questions, \eg ``\textit{The next step is to heat the mug}''. 
\textbf{Suggestions} offer additional information and suggest extra actions to rectify previous errors, \eg ``\textit{You can heat the mug after adding sugar and mix it again},'' where the user forgot to add sugar.
\textbf{Interventions} inform a user of irreparable situations and recommend starting over from an earlier point, \eg ``\textit{You should start over with filing the mug with milk instead of water},'' where the user mistakenly filled the mug with water.

\subsection{Statistics}
\label{ssec:stats}

\begin{figure}[t]
    \centering
    \includegraphics[width=\linewidth]{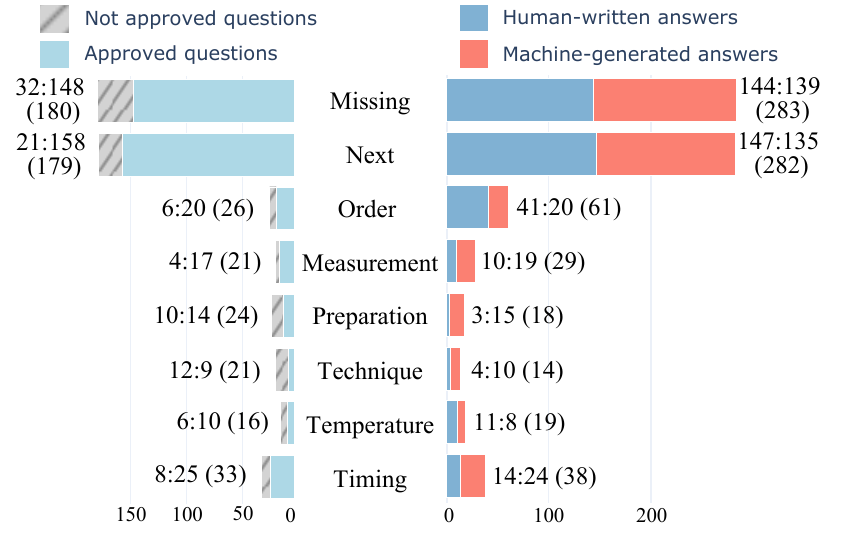}
    \caption{Question approval counts (left) and the answer counts by source (right) for each question type.}
    \label{fig:question-answer-stat}
\end{figure}


\begin{table}[!t]
\centering
\fontsize{8}{9}\selectfont
\begin{tabular}{c c c}
    \toprule
    & Ours & Human (est.) \\
    \midrule
    Cost / Hour & 5 USD / 0.5 Hour & 800 USD / 40 Hour \\
    \bottomrule
\end{tabular}
\caption{Cost comparison between our human-LLM collaborative approach and a full-human approach for generating 500 QA pairs. For the latter, we asked one annotator to create 50 QAs from scratch, which took 4 hours at an assumed hourly rate of 20 USD.}
\label{tab:cost}
\end{table}

\begin{figure}[t]
    \centering
    \includegraphics[width=\linewidth]{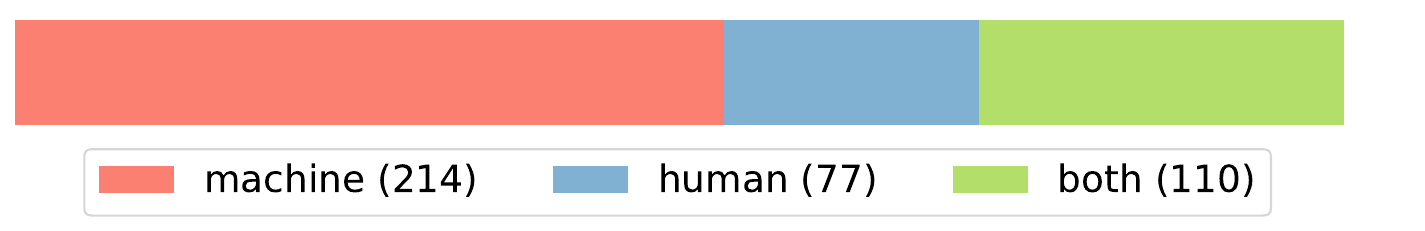}
    \caption{Answer source: The number of examples with only machine-generated answers, only human-written answers, or both types of answers (count).}
    \label{fig:answer-source-ratio}
\end{figure}
\begin{figure}[t]
    \centering
    \includegraphics[width=\linewidth]{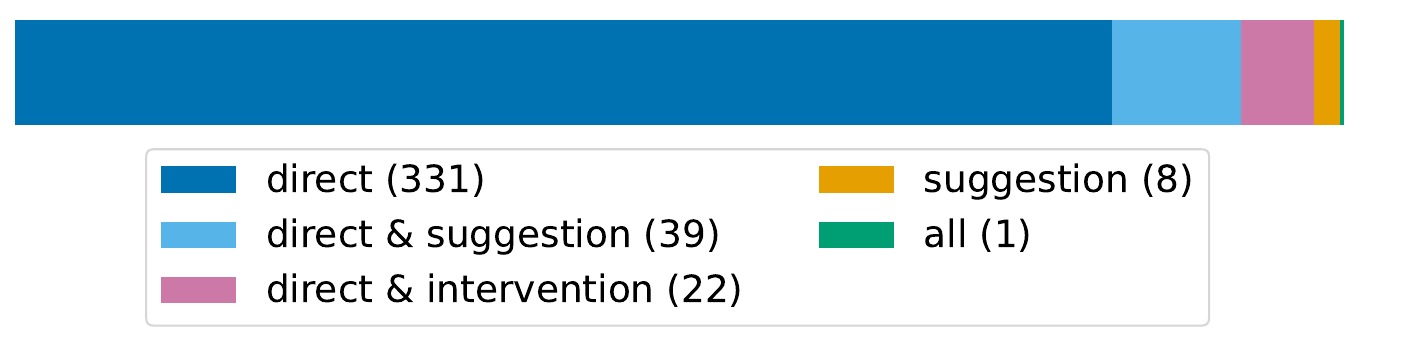}
    \caption{Answer type: The number of examples with only direct answers, direct answers and suggestions, direct answers and interventions, only suggestions, or all types of answers (count). Note that other combinations, i.e., only interventions or suggestions and interventions, are not found in our dataset.}
    \label{fig:answer-type-ratio}
\end{figure}

We show the general statistics of our dataset in~\autoref{tab:general-stats}. 
Among the 401 examples, 225 examples have no errors in previous steps (clean) and 176 examples have at least one error in previous steps.
\autoref{fig:question-answer-stat} illustrates the high approval rate for questions, while approximately 50\% of answers were added by humans through the verification process. 
In addition to showing the total count of each answer characteristic, we also count the number of examples with each combination of answer sources and types, as shown in~\autoref{fig:answer-source-ratio} and~\ref{fig:answer-type-ratio}.
For these analyses, while we used the answer source information retained through the annotation process, we obtained the answer type information by asking one annotator to categorize each answer into one of three types.
We further compare the cost of our human-LLM collaborative approach and the estimate of the full-human annotation in~\autoref{tab:cost}.
QA annotation for evaluation/test data typically consists of two steps: initial QA creation, followed by verification to assure the quality. 
We only compare the cost of the QA creation/generation part as our annotation framework replaces humans with LLMs in the initial QA creation (\S\ref{sec:annotation}).
According to the table, our approach substantially reduces the cost of the QA creation part.

\section{Annotation: \textit{Generate-then-Verify}}
\label{sec:annotation}

In this work, we take a human-LLM collaborative approach to annotate QA pairs: 
LLMs \textit{generate} QA pairs with lower cost, \textit{then} humans \textit{verify} them to ensure quality.
We hypothesize that LLMs can substantially generate valid questions when given sufficient information, inspired by synthetic data generation~\citep{mangalam2023egoschema, wu-etal-2024-synthetic}.
Specifically, we leverage existing annotations of action and error labels to form textual prompts.
We note that, as our annotation framework is LLM agnostic, it can plug and play LLMs, and importantly, it can benefit from ongoing LLMs' improvement. 

\subsection{Source \& Preprocess}
\label{ssec:preprocess}
We chose CaptainCook4D~\citep{peddi2023captaincook4d} as our data source because it includes explicit instructions and user recordings with human-annotated actions and error labels. 

First, we extract video segments of various lengths using annotated action temporal segmentations.
Given an original recording $video_{original}$ with $n$ actions, we create $n+1$ video clips $video_{0:k}$ such that each clip contains the first $k$ recording steps $S^{video}_{0:k} = \{s^{video}_0, ..., s^{video}_{k}\}$ ($k = 0, 1, ..., n$ and $s^{video}_0 = \varnothing$).
Each video clip with its corresponding recipe constitutes one data example $d_{init}$: 
$d_{init} = \langle recipe, video_{0:k} \rangle $.
From each $d_{init}$, we augment 2\textasciitilde{}8 examples by adding each question type based on existing error labels: 
$d_{type} = \langle recipe, video_{0:k}, type \rangle$.
Specifically, we create $d_{type}$ for each, \textit{next} and \textit{missing}.
For other six types, we create $d_{type}$ only when the last recording step $s^{video}_k$ in $video_{0:k}$ has a corresponding error annotation. 
This is based on our preliminary experiment, revealing that LLMs struggled to generate those six types of questions when no corresponding errors were annotated.

After obtaining approximately 11,000 examples from this process, we sampled 500 examples by taking the following points into account to increase diversity:
(1) Sample one example for each question type from each recording; 
(2) Evenly sample examples with errors (\textit{noisy}) and without errors (\textit{clean}) in previous steps for all types;
(3) Evenly sample examples that do and do not have target recipe steps for \textit{next} and \textit{missing} types.\footnote{Example question without target recipe steps: ``Did I miss any step?'' ``No.''}
Note that the activities in CaptainCook4D do not always result in the expected outcomes, i.e., failed procedures are included. 
Hence, our \textit{noisy} examples allow us to generate QAs on top of unaddressed and/or irreparable errors from previous steps.

\subsection{QA Generation}
\label{ssec:qa-generation}
\begin{figure}[t]
    \centering
    \includegraphics[width=0.90\linewidth]{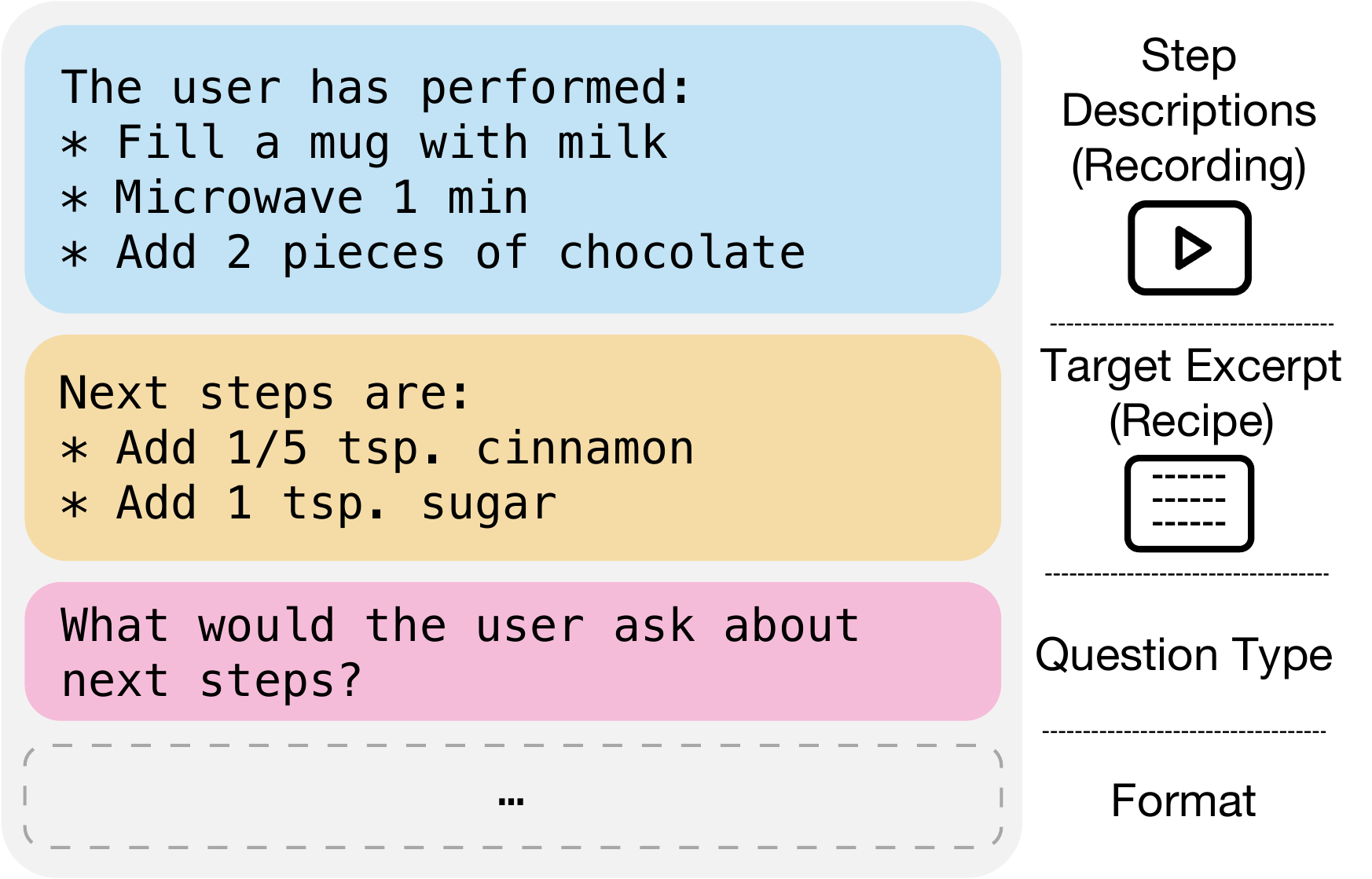}
    \caption{Example prompt with recording steps to embed recording information, an on-target excerpt from a recipe, and a question type for QA generation.}
    \label{fig:prompt-qa-generation}
\end{figure}

Given $d_{type}$, we prompt an LLM to generate a QA pair. 
\autoref{fig:prompt-qa-generation} shows a shortened example of our prompt, and an actual example is available in Appx.~\ref{appx:qa-generation}.
Each prompt consists of three pieces of information: 
(1) the textual description of $S^{video}_{0:k}$, (2) an excerpt from $recipe$ to embed what is next, missing, or incorrect, and (3) $type$, the question type to guide generation.
We feed the prompts to an LLM to generate $l$ QA pairs, from which we randomly pick one ($l = 3$). 
This is based on our preliminary experiment, where single pair generation often leads to monotonic question expression, \eg ``What is the next step?'' across multiple $next$ examples.
With GPT-4o as our QA generator, we obtain 500 examples with a pair of a machine-generated question $q^m$ and its machine-generated answers $A^m = \{a_1^m, a_2^m, ...\}$:
$d_{gen} = \langle recipe, video_{0:k}, q^m, A^m, \rangle$.

In fact, it is not trivial how to represent information in prompts and which LLMs to use to obtain better QA pairs.
We conduct ablation studies to determine the prompt template and LLM.

\begin{table}[!t]
\fontsize{8}{8}\selectfont
\centering
\begin{tabular}{c|ccc}
    \toprule
    \multirow{3}{*}{Recording} & \multicolumn{3}{c}{Recipe} \\ \cline{2-4}
    \\[-0.7em]
     & DOT & image & excerpt \\ \midrule
    frames & 0.43 & 0.53 & 0.65 \\
    step & 0.54 & 0.60 & 0.71 \\ \bottomrule
\end{tabular}
\caption{Approval rate comparison for QA generation prompts with a fixed LLM, GPT-4o.}
\label{tab:prompt-explore-qa-generation}
\end{table}
\begin{table}[!t]
\setlength{\tabcolsep}{4pt}
\fontsize{8}{9}\selectfont
\centering
\begin{tabular}{c | c c c}
    \toprule
    \multirow{3}{*}{Template} & \multicolumn{3}{c}{QA Generator} \\
    \cline{2-4}
    \\[-0.75em]
    & GPT-4o & \makecell[c]{Gemini 1.5 pro} & \makecell[c]{Claude 3.5 Sonnet} \\
    \midrule
    excerpt \& step & 0.71 & 0.69 & 0.68 \\
    \bottomrule
\end{tabular}
\caption{Approval rate comparison for QA generators.}
\label{tab:model-selection-qa-generation}
\end{table}

\paragraph{Prompt Exploration}
In this ablation study, we compare the methods to embed $recipe$ and $video$ information, using small samples of $d_{type}$ and a fixed QA generator.
In a typical full-human annotation scenario, $recipe$ and $video_{0:k}$ are represented as a whole recipe and a video segment, respectively.
Inspired by this, we consider the following settings:
For a recipe, we compare three methods: a whole recipe as a DOT language graph~\citep{koutsofios1991drawing} (``DOT''), a whole recipe graph as an image (``image''), and only the on-target excerpt from a recipe (``excerpt'').
We use DOT to accurately represent the partial graph information in a recipe.
For a video segment, we feed a video segment as sampled frames (``frame'') and a list of step descriptions (``step''). 
Actual example prompts are available in Appx~\ref{appx:qa-generation}.
We generate 80 questions for each combination using GPT-4o and ask one annotator to check if they are multimodal procedural questions. 
\autoref{tab:prompt-explore-qa-generation} shows the approval rate for each combination, \ie how many generated questions passed the check.
We found that feeding the combination of the excerpt from a recipe and step descriptions resulted in the most approved QA pairs.

\paragraph{QA Generator Selection}
In the second ablation study, we compare QA generators by fixing the prompt template (excerpt \& step).
The following LLMs are our candidates: GPT-4o, Claude 3.5 Sonnet, and Gemini 1.5 Pro~\citep{team2024gemini}.\footnote{These model versions were used throughout the paper: \texttt{gpt-4o-2024-08-06}, \texttt{claude-3-5-sonnet-20240620}, \texttt{gemini-1.5-pro-001}.} 
Similar to the prompt exploration, we use the approval rate as our metric based on the annotator's judgments.
As shown in~\autoref{tab:model-selection-qa-generation}, the performance is not very different, yet, we found that GPT-4o generates slightly more valid questions.

\subsection{Verification}
\label{ssec:verify}
LLM-generated questions and answers are not guaranteed to be valid.
Thus, we resort to human annotators to ensure the quality of our evaluation data.

\paragraph{Criteria}
For questions, annotators check if each is a valid multimodal procedural question, as described in \S\ref{ssec:multimodal-procedural-qa}, and assess for naturalness, clarity, and grammatical correctness.
For answers, annotators verify the correctness of each answer.

\paragraph{Process}
Our verification process involves two stages: 
In the first stage, two annotators independently verify each question and its answers in $d_{gen}$.
When a question is marked as valid, its answers are shown to annotators to verify. 
Otherwise, annotators move on to the next example.
During answer verification, annotators can add human-written answers $A^h = \{a^h_1, a^h_2, ...\}$, including suggestions and interventions, when any generated answers are incorrect or additional correct answers are missing.
When two annotations for one $d_{gen}$ conflicts or at least single $a^h$ exists, an additional annotator (\ie adjudicator) further verifies examples to make the final judgment or to have an additional check.
More details are available in Appx~\ref{appx:verification}.

We first created an annotation guideline and hired 6 people with graduate degrees in NLP-related fields for our annotation.
Among the participants, five people served as first-stage annotators, while the other, who was also involved in the guideline development, took the adjudicator's role.
This adjudicator performed the answer categorization, verification, and human judgment in~\S\ref{ssec:stats}, \S\ref{ssec:qa-generation}, and \S\ref{ssec:llm_as_a_judge} as well.
To help familiarize the annotators with the task, we conducted a training phase in which each annotator verified 20 examples and received personalized feedback. 
Following the training session, we initiated the main phase.
On average, judgment agreements were $0.87$ for both questions and answers.
After the verification, we obtained 401 verified examples $d_{ver} = \langle recipe, video_{0:k}, q^m_{ver}, A^m_{ver}, A^h_{ver} \rangle$. 

\section{Benchmarking}
\label{sec:benchmark}

On our \dataset{}, we provide the baseline results of existing models to facilitate the development of a user-support system for procedural activities.
Considering that our task contains natural language answers, we employ an LLM-based metric to evaluate the performance of the baselines.

\subsection{Target Models}
\label{ssec:target_model}
We consider the following approaches:

\paragraph{Unimodal Model}
One baseline consists of a text-only unimodal model, which shows how many examples in \dataset{} can be solved/guessed solely from textual information (\ie instructions and questions). 
Vision-only unimodal models are not considered, as inputs without questions would not guide the model to generate on-target answers.
We employ Llama 3.1 Instruct~\citep{dubey2024llama}.

\paragraph{Socratic Model}
Another baseline is a two-model pipeline: one generates captions from visual inputs, and the other generates answers based on those captions and text information. 
This approach demonstrates how many questions can be answered with restricted cross-modal/frame reasoning.
We use LLaVA 1.5~\citep{Liu_2024_CVPR} for image captioning and Llama 3.1 Instruct for text-based reasoning. 

\paragraph{Multimodal Model}
As one of our main targets, we assess open multimodal models, especially the ones tailored towards video understanding. 
Based on the strong performance on the existing multimodal benchmarks, \eg MMMU~\citep{Yue_2024_CVPR} and Video-MME~\citep{fu2024video}, we evaluate VideoLLaMA2~\citep{damonlpsg2024videollama2} and Qwen2-VL~\citep{wang2024Qwen2VL}.
Finally, we test proprietary multimodal models (\ie GPT-4o, Claude 3.5 Sonnet, and Gemini 1.5 Pro) considering their strong performance in various benchmarks.

\begin{figure}[t]
    \centering
    \includegraphics[width=0.90\linewidth]{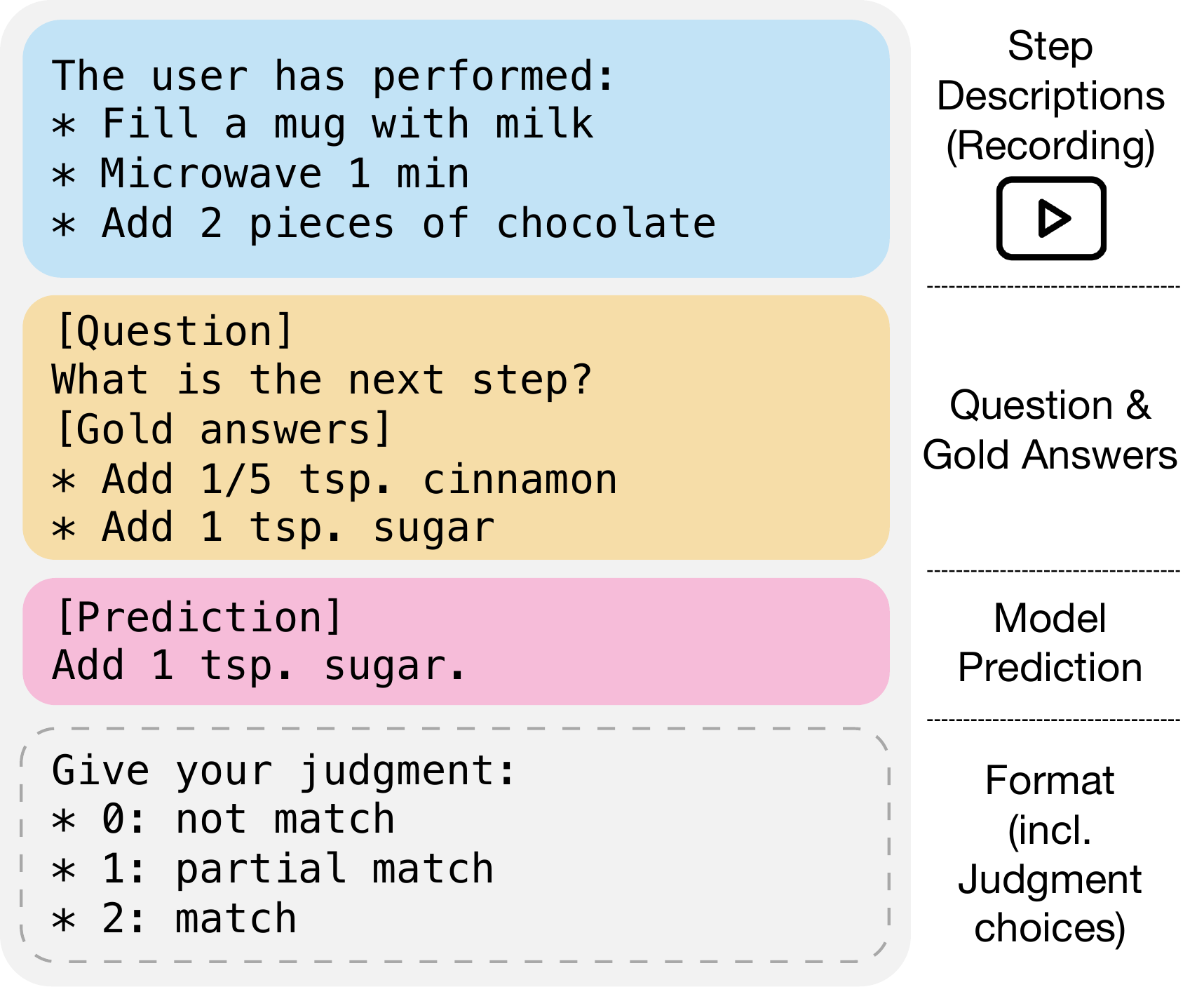}
    \caption{Example prompt for LLM Scoring with recordings as context information, a question with its gold answer(s), and a model prediction.}
    \label{fig:prompt-eval}
\end{figure}

\begin{table}[!t]
\fontsize{8}{9}\selectfont
\setlength{\tabcolsep}{10pt}
\centering
\begin{tabular}{c|ccc}
    \toprule
    \multirow{3}{*}{\#choice} & \multicolumn{3}{c}{Context information} \\ \cline{2-4}
    \\[-0.7em]
     & default & DOT & step \\ \midrule
    binary & 0.67/0.86 & 0.57/0.80 & 0.71/0.89 \\
    ternary & 0.67/0.69 & 0.58/0.64 & 0.76/0.75 \\ \bottomrule
\end{tabular}
\caption{LLM-based scoring prompt comparison (Pearson/Acc.)}
\label{tab:prompt-explore-eval}
\end{table}
\begin{table}[!t]
\fontsize{8}{9}\selectfont
\setlength{\tabcolsep}{4pt}
\centering
\begin{tabular}{c | c c c}
    \toprule
    \multirow{3}{*}{Template} & \multicolumn{3}{c}{Evaluator} \\ \cline{2-4}
    \\[-0.7em]
    & GPT-4o & \makecell[c]{Claude 3.5 Sonnet} & \makecell[c]{Gemini 1.5 Pro} \\
    \midrule
    ternary \& step & 0.83/0.82 & 0.79/0.77 & 0.66/0.68 \\
    \bottomrule
\end{tabular}
\caption{Evaluator comparison (Pearson/Acc.)}
\label{tab:model-selection-eval}
\end{table}
\begin{table*}[!t]
\fontsize{8}{8}\selectfont
\setlength{\tabcolsep}{3pt}
\centering
\begin{tabular}{l c c c c c c c c c c c}
    \toprule
    \multirow{2}{*}{Model} & \multirow{2}{*}{Avg.} & \multicolumn{2}{c}{Error} & \multicolumn{8}{c}{Question Type} \\
    \cmidrule(lr){3-4} \cmidrule(lr){5-12}
    & & clean & noisy & missing & next & order & measurement & preparation & technique & temperature & timing\\
    \midrule
    Llama 3.1 70B Instruct & 31.5 & 33.2 & 30.2 & 35.5 & 38.3 & 12.5 & 2.9 & 14.3 & 5.6 & 30.0 & 20.0 \\
    \midrule
     \makecell[l]{LLaVA 1.5 13B (50f, 288p) \\ \& Llama 3.1 70B Instruct} & 37.5 & 38.9 & 36.4 & 41.9 & 45.3 & 12.5 & 11.8 & 14.3 & 11.1 & 50.0 & 18.0 \\
    \midrule\
    VideoLLaMA2 72B (8f, 336p) & 39.8 & 49.4 & 32.2 & 46.3 & 49.7 & 20.0 & 0.0 & 21.4 & 11.1 & 25.0 & 8.0 \\
    Qwen2-VL 72B (100f, 336p) & 31.2 & 32.1 & 30.4 & 34.1 & 37.0 & 45.0 & 0.0 & 10.7 & 0.0 & 20.0 & 14.0 \\
    \midrule
    GPT-4o (50f, 765p) & 40.4 & 39.5 & 41.1 & 39.9 & 45.9 & 45.0 & 29.4 & 17.9 & 27.8 & 55.0 & 24.0 \\
    Gemini 1.5 Pro (50f, 765p) & 25.2 & 27.0 & 23.8 & 27.4 & 29.7 & 15.0 & 17.6 & 7.1 & 16.7 & 20.0 & 12.0 \\
    Claude 3.5 Sonnet (10f, 765p) & 44.1 & 48.9 & 40.4 & 44.6 & 58.2 & 27.5 & 8.8 & 14.3 & 5.6 & 25.0 & 28.0 \\

    \midrule
    Human* & (74.5) & (83.5) & (65.8) & --- & --- & --- & --- & --- & --- & --- & --- \\
    \bottomrule
\end{tabular}
\caption{Benchmark result: The average of all the examples, the averages of examples with (noisy) and without errors (clean) in previous steps, and the averages for the same question-type examples. f and p denote the number of frames and image resolution used for each model. Note that each category contains a different number of examples. *: Human performance is based on the sampled 20 examples. 
}
\label{tab:benchmark-result}
\end{table*}
\begin{table*}[!t]
\fontsize{8}{8}\selectfont
\setlength{\tabcolsep}{3pt}
\centering
\begin{tabular}{l c c c c c c c c c}
    \toprule
    \multirow{3}{*}{Model} & \multirow{3}{*}{Avg.} & \multicolumn{3}{c}{Answer Source } & \multicolumn{5}{c}{Answer Type} \\
    \cmidrule(lr){3-5} \cmidrule(lr){6-10}
    & & machine & human & both & direct & direct \& suggestion & direct \& intervention & suggestion & all\\
    \midrule
    Llama 3.1 70B Instruct & 31.5 & 33.4 & 28.0 & 30.5 & 31.7 & 30.8 & 34.1 & 25.0 & 0.0 \\ 
    \midrule
    \makecell[l]{LLaVA 1.5 13B (50f, 288p) \\ \& Llama 3.1 70B Instruct} & 37.5 & 40.7 & 32.9 & 34.8 & 38.8 & 34.6 & 34.1 & 12.5 & 0.0 \\ 
    \midrule
    VideoLLaMA2 72B (8f, 336p) & 39.8 & 45.1 & 32.9 & 34.3 & 41.7 & 34.6 & 31.8 & 12.5 & 0.0 \\ 
    Qwen2-VL 72B (100f, 336p) & 31.2 & 35.0 & 22.0 & 30.5 & 30.8 & 33.3 & 27.3 & 37.5 & 100.0 \\ 
    \midrule
    GPT-4o (50f, 765p) & 40.4 & 41.4 & 29.3 & 47.1 & 40.2 & 39.7 & 52.3 & 18.7 & 50.0 \\ 
    Gemini 1.5 Pro (50f, 765p) & 25.2 & 27.1 & 18.3 & 26.7 & 24.8 & 25.6 & 34.1 & 12.5 & 50.0 \\ 
    Claude 3.5 Sonnet (10f, 765p) & 44.1 & 48.1 & 35.4 & 42.9 & 45.0 & 42.3 & 43.2 & 25.0 & 0.0 \\ 
    \bottomrule
\end{tabular}
\caption{Answer-focused benchmark result breakdown: The average of all the examples, the averages of examples with only machine-generated answer(s), human-written answer(s), and both; The averages of examples with only direct answer(s), direct and suggestion(s), direct and intervention(s), only suggestion(s), and all answer-types. 
f and p denote the number of frames and image resolution used for each model.
Note that each category contains a different number of examples. 
}
\label{tab:benchmark-result-answer}
\end{table*}

\subsection{LLM-as-a-Judge}
\label{ssec:llm_as_a_judge}
Evaluating natural language itself is a challenging task due to multiple correct answers and their possible variations for the same question. 
In place of string-based metrics, \eg BLEU~\citep{papineni-etal-2002-bleu}, which often struggle with such an answer diversity, LLM-based metrics, i.e., LLM-as-a-judge~\citep{NEURIPS2023_91f18a12} are getting increasing attention. 
Considering possible correct answers, we also employ LLM-as-a-judge in the experiment.
\autoref{fig:prompt-eval} shows our shortened prompt for our LLM-based scoring.
As a calibration process, we conduct ablation studies to choose which information to feed in prompts and an LLM as our evaluator.

\paragraph{Prompt Exploration and Evaluator Selection}
\label{ssec:prompt_eval}
We aim to identify a prompt template and an LLM that yields a high correlation with human judges.
We consider two key aspects in templates: 
1) the number of choices in the Likert scale and 2) the context information.
For choices, we consider ``binary'' (\textit{match} and \textit{unmatch}) and ``ternary'' (\textit{match}, \textit{partial-match}, and \textit{unmatch}).
For context, we examine three settings: 
With a question, gold answers, and a predicted answer as the fundamental elements (``default''), we then incorporate either instruction (``DOT'') or step descriptions from recordings (``step''). 
Candidate evaluators include GPT-4o, Claude 3.5 Sonnet, and Gemini 1.5 Pro.
In the experiment, we feed inputs based on the verified examples in~\S\ref{ssec:qa-generation} to LLM-evaluators to obtain predictions.
Then, we obtain judgments from these LLMs with all combinations.
As a comparison, we obtain human judgments, where one person judges the predictions with both binary or ternary options. 
We consider Pearson correlation coefficient~\citep{pearson1895vii} and match accuracy as our metrics.
\autoref{tab:prompt-explore-eval} shows the average scores across three evaluator models. 
We found that the combination of ``ternary'' and ``step'' produces the highest correlation.
With the best combination, we compare the evaluators. 
\autoref{tab:model-selection-eval} shows that GPT-4o has the best correlation with the human judgments.
In the benchmark experiment, we scaled judgment scores from 0-2 to 0-100 by multiplying 50.

\subsection{Results}
\label{ssec:result_benchmark}
In this experiment, we also obtained human performance as a comparison. 
We asked five first-stage annotators (\S\ref{ssec:verify}) to solve 20 samples, out of 401 total examples, which they had not previously checked during the verification process. 
The sampling was done due to our budget. 
Performers were provided only recipes and video segments with questions.
In~\autoref{tab:benchmark-result}, we provide the average performance, as well as the breakdown based on previous step types and question types.
It shows that all the models we benchmark lag behind human performance, even the competitive proprietary models. 
Among the models, Claude 3.5 Sonnet performs relatively better than others, although the differences are somewhat marginal.
In general, \textit{clean} examples are easier for models than \textit{noisy} examples, although the gap varies depending on each model.
Proprietary models are, on average, better on step-specific questions.
Additionally, we show the breakdown based on answer sources and types in~\autoref{tab:benchmark-result-answer}.
From the table, we can see that models generally perform better on examples with only machine-generated answers, although each model exhibits different preferences.
Furthermore, we investigate the effect of answer counts of examples on performance.
There is a weak common trend that models perform well on examples with a single answer and with 4 answers. 
Considering our results do not always align with those in the public benchmarks like Video-MME, we believe our \dataset{} can be complementary in evaluating models' multimodal capabilities. 

\begin{figure}[!t]
    \centering
    \includegraphics[width=\linewidth]{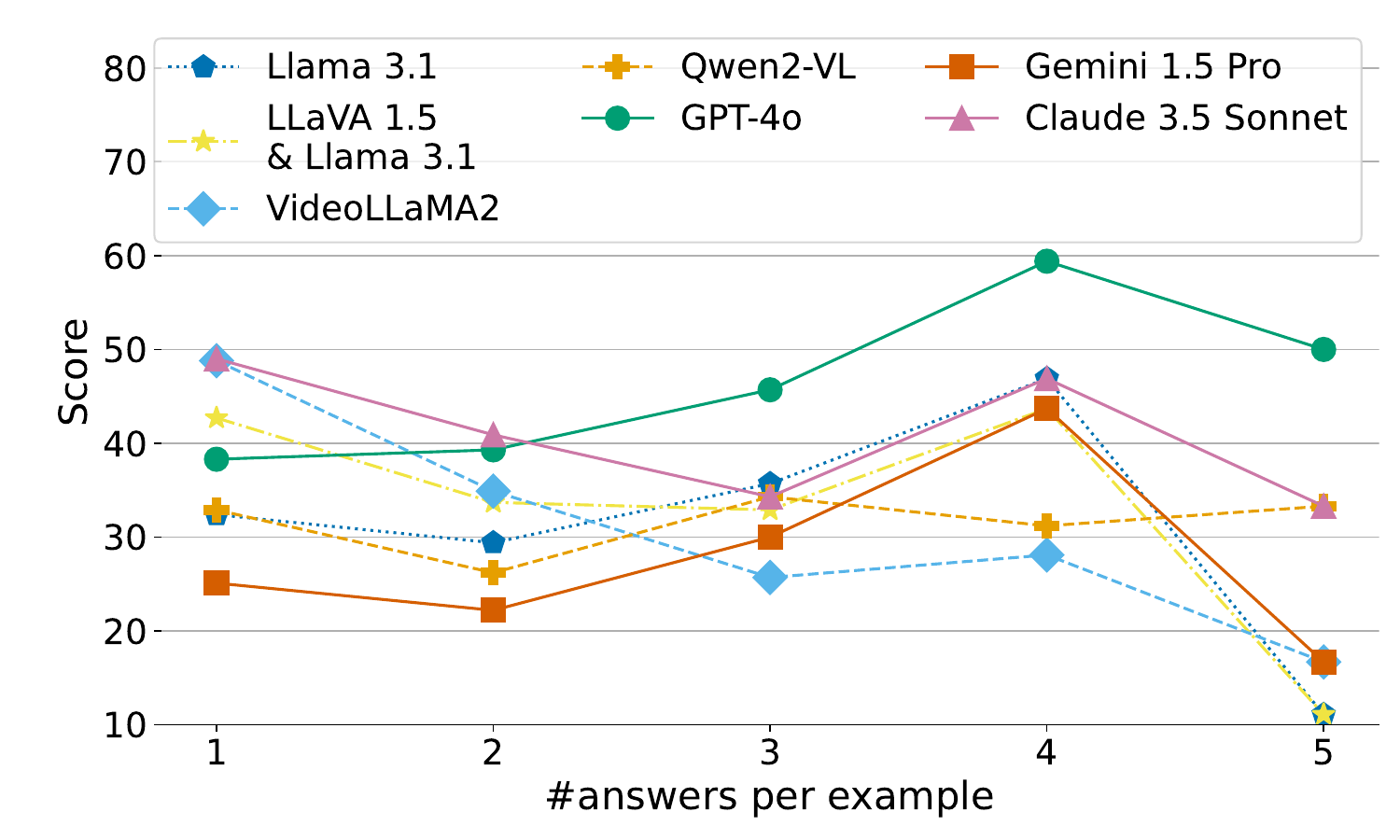}
    \caption{Model performance with different numbers of answers. Note that each category contains a different number of examples and examples with more than 5 answers are excluded due to their small counts.}
    \label{fig:answer-num}
\end{figure}

\section{Self-Preference Bias Analysis}
\label{sec:bias}


\begin{table}[!t]
\fontsize{8}{9}\selectfont
\centering
\begin{tabular}{c|cc|cc}
    \toprule
    \multirow{2}{*}{Predictor} & \multicolumn{2}{c|}{Generator} & \multicolumn{2}{c}{Evaluator} \\
    & GPT-4o & Gemini & GPT-4o & Gemini \\ \midrule
    GPT-4o & 22.8 & 36.4 & 27.5 & 31.4 \\
    Gemini & 14.0 & 12.7 & 21.6 & 24.5 \\
    \bottomrule
\end{tabular}
\caption{Result of generator-predictor and predictor-evaluator self-preference bias checks. Each number represents the score by a human evaluator. Gemini denotes Gemini 1.5 Pro.}
\label{tab:bias-check}
\end{table}

Prior studies report that LLMs may introduce self-preference bias: 
``an LLM favors its own outputs over texts from other LLMs and humans.''~\citep{panickssery2024llm}
This can be a critical issue when LLMs play multiple roles, as in our experiment, i.e., use LLM-generated QAs to evaluate LLMs with LLM-as-a-judge.
To justify the fairness of \dataset{} as a benchmark dataset, we investigate:
(1) \textbf{generator-predictor} self-preference bias, where the generator's outputs harbor styles or characteristics that make it easier for the model to answer, and (2) \textbf{predictor-evaluator} self-preference bias, where the evaluator favors their own outputs, rather than objectively assessing the accuracy or quality of the predictions.
Our experiments show no noticeable sign of biases. 

\subsection{Bias: Generator-Predictor}
\label{ssec:bias_gen}
We investigate if questions generated by an LLM is easier for the same LLM to derive answers.
To conduct a control experiment, we change generators and predictors, while fixing other variables, i.e., the verification person and evaluator (manual).
According to~\autoref{tab:bias-check}, we did not find an indication that a model scores higher on its generated questions.
One reason could be the modality difference between generation (text-only) and prediction (text \& visual inputs), but we leave it for future work.

\subsection{Bias: Predictor-Evaluator}
\label{ssec:bias_eval}
We then examine the original self-preference bias, i.e., if an LLM favors their own predictions over others.
We fix generators (\ie verified QAs from three LLMs), and change predictors and evaluators with the same set of LLMs for each.
Contrary to the previous work, ~\autoref{tab:bias-check} shows no sign of the bias.
We believe that QA evaluation is more objective than the summarization used by~\citet{panickssery2024llm}, resulting in less room for model-based bias. 
We again put deeper analysis as future work.

\section{Related Work}
\label{sec:related_work}

\paragraph{Procedural Activity Understanding}
\label{ssec:related_work_proc}
The research community constructed various datasets to improve the machine's understanding of procedural activities in videos: 
Breakfast~\citep{Kuehne_2014_CVPR}, YouCook2~\citep{Zhou_Xu_Corso_2018}, COIN~\citep{yansong2019coin}, Assembly101~\citep{sener2022assembly101}, and CaptainCook4D~\citep{peddi2023captaincook4d}, to name a few.
With those datasets, models are typically evaluated on tasks like action recognition and temporal action localization, framed as classification tasks. 
In this work, we propose QA as the formulation, which aligns better with real-world scenarios.

\paragraph{Video QA Dataset}
\label{ssec:related_work_vqa}
QA as a task formulation is increasingly adopted for video QA datasets, e.g., NExT-QA~\citep{xiao2021next}, EgoSchema~\citep{mangalam2023egoschema}, OpenEQA~\citep{majumdar2024openeqa}, Video-MME~\citep{fu2024video}, \textit{inter alia}.
While they are multimodal, i.e., a model takes video frames and a textual question as inputs, we argue that they are still rather video-oriented as only a short question consists of the textual part, compared to a pile of images from a video. 
While GazeVQA~\citep{ilaslan-etal-2023-gazevqa} uniquely focuses on procedural tasks as QA, instructions are yet explicitly provided to systems, hence, only a short question with multiple choices and a video are the inputs. 
For enhanced cross-modal comprehension, we present \dataset{} where textual instructions are necessary to derive a correct answer in addition to a video and question (\S\ref{ssec:task}).

\paragraph{Synthetic Evaluation Data}
\label{ssec:related_work_synthetic}
Along with the advancement of LLMs, synthetic data generation is widely explored in various phases of model development, including pretraining~\citep{gunasekar2023textbooks, maini-etal-2024-rephrasing} and instruction tuning~\citep{wang-etal-2023-self-instruct, adler2024nemotron}. 
Compared to those phases, it is underexplored in generating evaluation data with LLMs~\citep{wu-etal-2024-synthetic}, possibly because of the following two reasons: 1) The quality assurance is lacking, which can be mitigated by introducing multi-step machine and manual curation steps as EgoSchema. 2) Potential biases may be introduced~\citep{zheng2024judging, panickssery2024llm}.
Addressing these challenges, we develop our \dataset{} with additional human checks (\S\ref{ssec:verify}), justified by the fairness-check experiments (\S\ref{sec:bias}). 

\section{Conclusion}
\label{sec:conclusion}
In this paper, we propose a human-LLM collaborative approach, \textit{Generate-then-Verify}, and develop a novel evaluation dataset, \dataset{}, for multimodal procedural activity understanding. 
\dataset{} consists of 401 QA pairs that require understanding both instructions and videos to derive answers, queried by questions. 
We also provide the baseline performance of existing models, showing that there is still a large gap in performance between humans and machines, even the competitive proprietary multimodal models.
We believe that \dataset{} can shed light on the new aspect of multimodal capabilities to facilitate model development. 
\newpage
\section{Limitation}
\label{sec:limitation}
We note a couple of limitations remain in this work.
First, the size of the dataset is relatively small. 
This may affect the confidence of performance comparisons when two models receive similar scores. 
We plan to increase the number of examples so that the research community can present their incremental progress, i.e., a few point improvements, with higher confidence~\citep{card-etal-2020-little}. 
However, despite its limited size., \dataset{} is carefully curated with a representative selection of questions and answers through our data annotation design. 
This enables it to serve as an effective testbed for multimodal foundation models for providing insights into model performance. 

Second, the domain is restricted to a single activity, cooking. 
Remember our annotation framework assumes the action and error labels, explicit instructions, and procedural videos. 
While our source dataset, CaptainCook4D, uniquely satisfies all the prerequisites, it does not apply to other existing datasets.
We leave it to future work how to extend our work to integrate other activities by making use of other datasets.

Third, the dataset is oriented toward English and Western countries, especially, the U.S.
CaptainCook4D contains recipes that originate from non-English speaking regions, e.g., ``Ramen'' or ``Bruschetta,'' but recipes and cooking environments are designed for people in the U.S.
We believe that our dataset can support the advancement of frontier multimodal models, which can also benefit diverse and/or general models. 
However, considering the ubiquitous potential of our target user-support systems, we hope to contribute to the development of systems for people in non-English, non-Western countries. 

Finally, we release our dataset as evaluation data, not for training data, which complies with the terms of use by OpenAI.\footnote{\url{https://openai.com/policies/row-terms-of-use/}}

\section{Ethical Consideration}
\label{sec:ethical_consideration}
In the dataset construction, we used LLMs that are pretrained on a massive web-scraped corpus, which may contain some toxic or biased information. 
We do not aim to include any prejudiced, offensive, or biased content in the dataset, and we did not find any in our verification process. 
CaptainCook4D received IRB approval and participants provided written consent in their data collection, and no private information included.\footnote{\url{https://github.com/CaptainCook4D/\#license--consent}}

\section*{Acknowledgements}
We thank the anonymous reviewers for their insightful comments. 
This work is partially supported by Programs for Bridging the gap between R\&D and the IDeal society (society 5.0) and Generating Economic and social value (BRIDGE) / Practical Global Research in the AI $\times$ Robotics Services, implemented by the Cabinet Office, Government of Japan.

\bibliography{anthology,custom}

\clearpage
\appendix

\section{\dataset{}}

\begin{figure}[h]
    \centering
    \includegraphics[width=\linewidth]{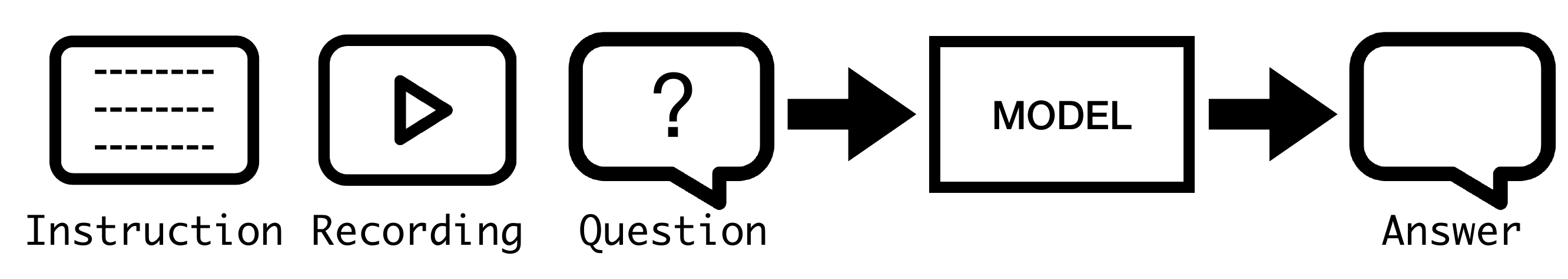}
    \caption{Task formulation of our dataset. Given recipe information, recording information, and a question, a model is to predict an answer. In our benchmark experiment, recipes and questions are fed as text, while recordings are passed as frames sampled from videos. Then, a model generates answers in text.}
    \label{fig:formulation}
\end{figure}

\autoref{fig:formulation} illustrates the task formulation.\footnote{Icons in~\autoref{fig:overview},~\ref{fig:prompt-qa-generation},~\ref{fig:prompt-eval}, and~\ref{fig:formulation} are from \url{https://www.flaticon.com}.}

\subsection{External Knowldge}
\label{appx:external-knowledge}
In \S\ref{ssec:multimodal-procedural-qa}, we define that our target multimodal procedural questions can be solvable from the combination of instruction and recording information.
Our task assumes the common sense that humans would have gained through their cooking experiences, in varying degrees. 
We note that this may introduce some ambiguity/subjectivity, regarding the boundary between common sense and external knowledge, as external knowledge is, to some extent, in the same spectrum as common sense. 
For instance, for well-experienced people, it can be too obvious (common sense) that replacing cilantro with parsley changes the flavor of a recipe, while others would think that is specialized/external knowledge.
To mitigate this subjectivity, we assign two annotators for each example in verification to account for this variance. 

\subsection{Other Verification Criteria}
\label{appx:other-criteria}
Additionally, we ask annotators to check the naturalness and clarity of questions. 
Naturalness is to check if a question is natural/makes sense to ask. 
For instance, when a question like ``Did I forget to do something before <stepX>?'' is asked, people usually assume that <stepX> has already been passed (with or without errors). 
So, if the question is asked when <stepX> is yet to be performed, this question will be unnatural/nonsensical. 
This criterion filters out this type of nonsensical question. 
Clarity filters out vague/too general questions, especially questions asking about non-procedural aspects. 
For instance, a question like ``What did I do wrong?'' can target non-procedural errors, e.g., ``Too many dishes are left in the sink.'' or ``The countertop is too messy.'' which we encountered in our preliminary QA generation and benchmarking experiments. 
To focus on the procedural questions, we added this criterion.

\subsection{Human-written QAs}
We obtain 50 \textit{next} questions by asking one of the annotators before conducting any verification process.
This provides the situation where one creates QA pairs without any prior knowledge about this work. 
They receive raw recipes and videos and create 50 $next$-type QAs from scratch, which took around 4 hours, as shown in~\autoref{tab:cost}.

\subsection{Additional Statistics}

\begin{figure}[t]
    \centering
    \includegraphics[width=0.8\linewidth]{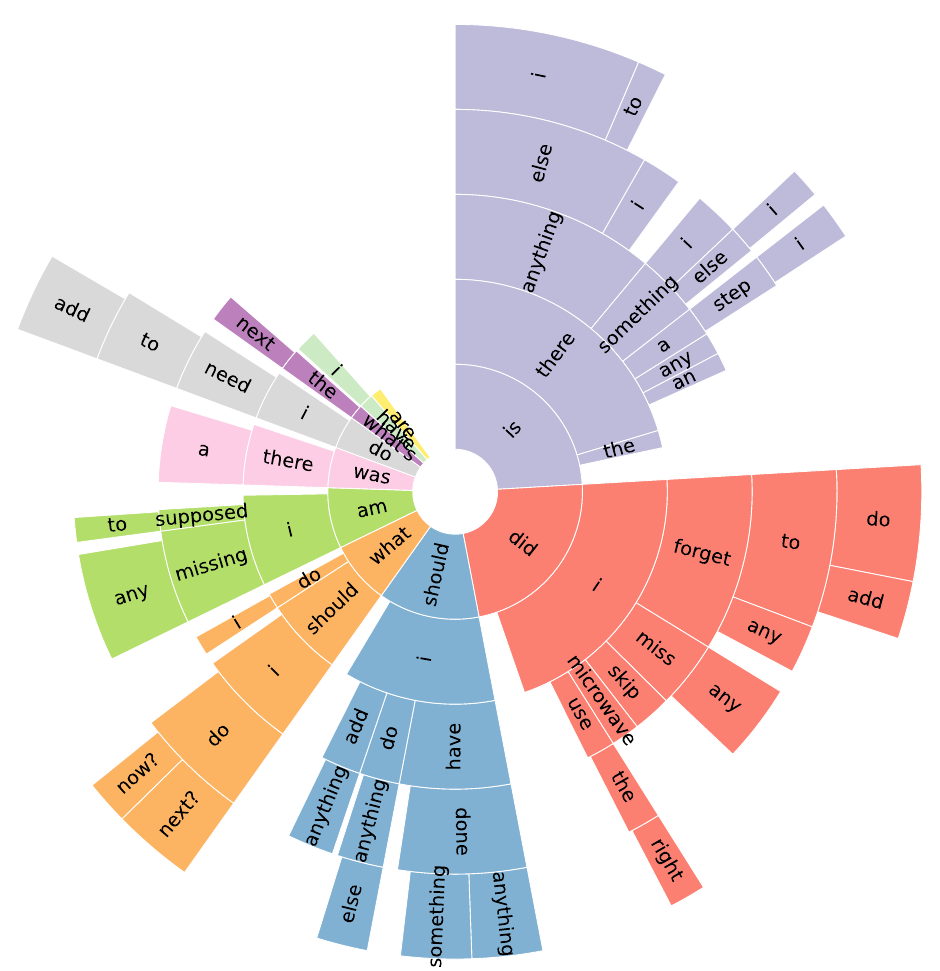}
    \caption{Count of 5 starting words in questions.}
    \label{fig:5gram-question}
\end{figure}

\begin{figure}[t]
    \centering
    \includegraphics[width=0.8\linewidth]{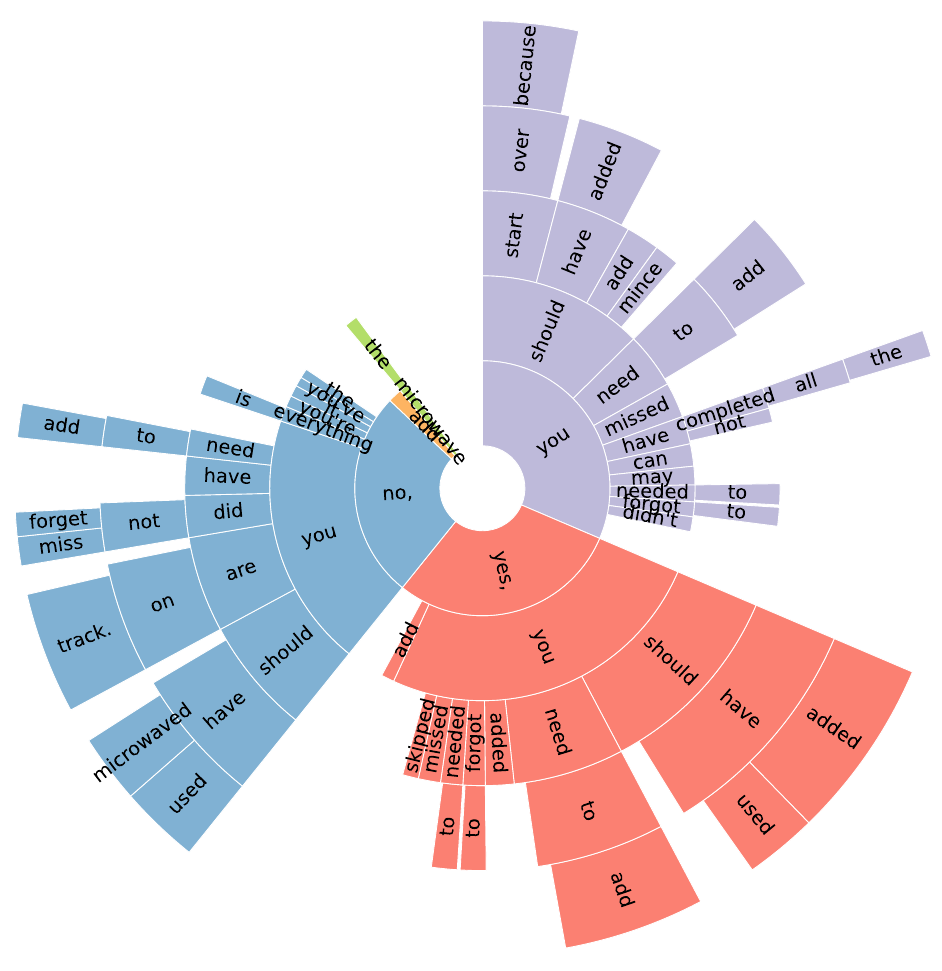}
    \caption{Count of 5 starting words in answers.}
    \label{fig:5gram-answer}
\end{figure}

In~\autoref{fig:5gram-question} and~\ref{fig:5gram-answer}, we show the counts of five starting words in questions and answers, sorted by question types.

In~\autoref{tab:cost}, we compare the cost between our approach and the full-human annotation approach. 
In addition, we compare machine-generated and human-written QAs in terms of question diversity using the type-token ratio, TTY ($ num\_unique\_words/total\_vocab$), and cosine similarity with E5 Small~\citep{wang2022text}. 
For human-written questions, we use the whole 50 questions to compute both numbers. 
For machine-generated questions, we sample 50 $next$ questions and compute the metrics. 
To reduce sampling variance, we sample 10 times and take the average of them.
TTY and cosine similarity are $0.07$ and $0.80$ for human-written questions and $0.09$ and $0.80$ for machine-generated questions. 
This suggests that both approaches produce similarly diverse questions at the surface and semantic level. 

\subsection{Statistical Power Analysis}
Following~\citet{card-etal-2020-little}, we conduct their statistical power analysis to estimate the performance difference required to detect statistical significance between systems with confidence. 
We first estimate the baseline accuracy based on the performance of GPT-4o, $0.4$ and the agreement rate based on  GPT-40 and Claude 3.5 Sonnet, $0.65$. 
Given these numbers, the simulation-based analysis\footnote{\url{https://github.com/dallascard/NLP-power-analysis}} shows that at least $8.5$ accuracy point difference would be needed to detect significance with $80\%$ confidence. 

\section{Annotation: \textit{Generate-then-Verify}}
\label{appx:annotation}

\begin{figure*}[t]
    \centering
    \includegraphics[width=\textwidth]{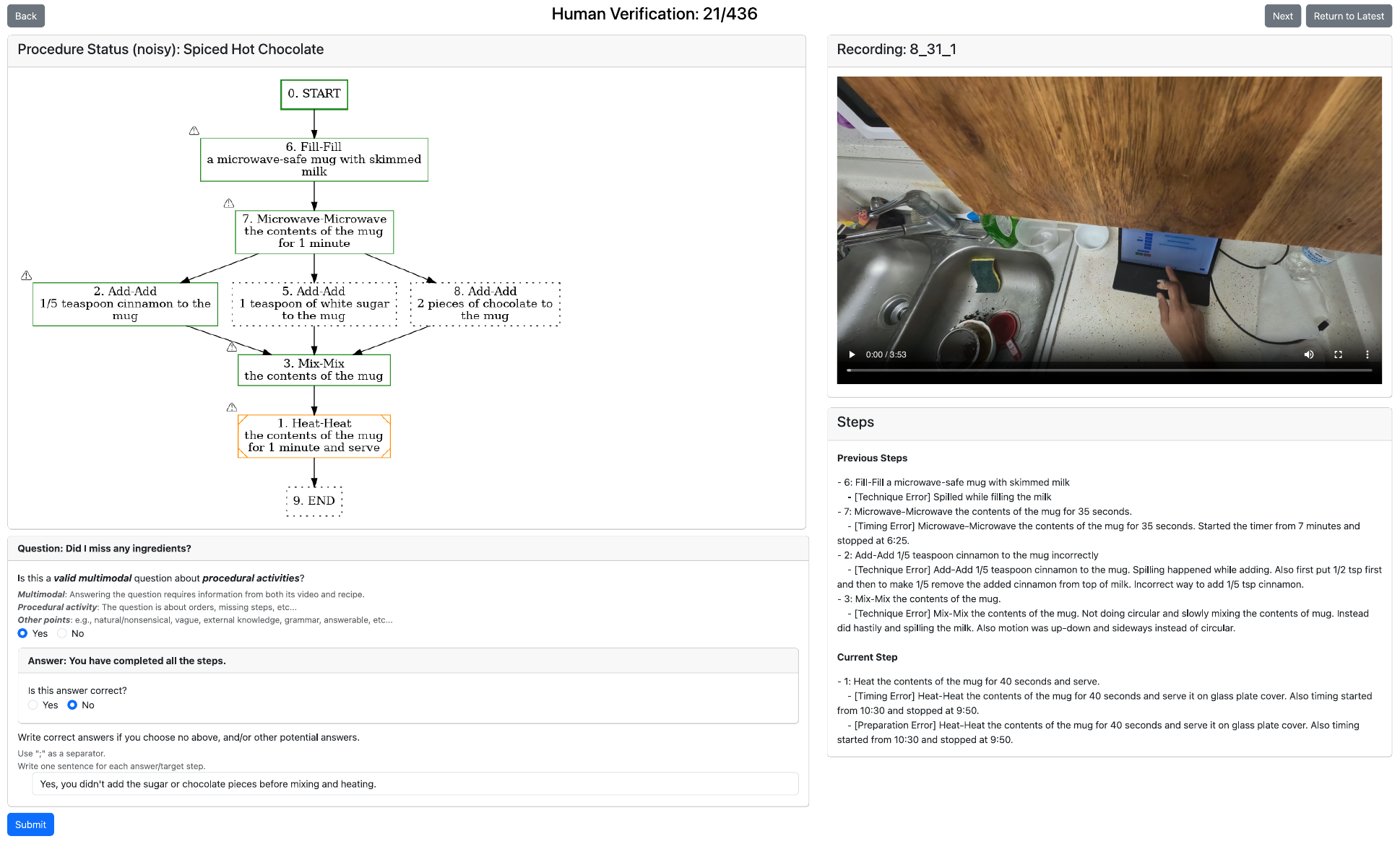}
    \caption{Verification interface.}
    \label{fig:interface}
\end{figure*}

\subsection{Preprocess}
Before the start of our automated preprocessing step, first, we corrected existing annotations in CaptainCook4D, especially about the orders, \eg by checking the consistency between the order and the timestamps.
In the preprocessing stage, we did not create $d_{type}$ of \textit{measurement}, \textit{preparation}, \textit{technique}, \textit{temperature}, \textit{timing}, and \textit{order} from $d_{init}$ when the last recording step did not have the corresponding error annotations. 
This is due to our preliminary experiments, where such cases tend to generate invalid multimodal procedural questions, i.e., the approval rate was much lower than others. 
This may be because not all actions can be associated with each type of question.
For instance, it is harder to create a sensical \textit{temperature} question from a step, ``Peel an onion.''
In addition, we skip creating $d_{type}$ in the following cases:
$video_{0:k}$ is too short, i.e., less than five seconds, which occasionally happens in the case of $video_0$; 
$missing$ questions for $video_0$;
The durations of $S_{0:k}^{video}$ overlaps with $s_{k+1}^{video}$, as it introduce extra step information in $video_{0:k}$.
Also, as for $S_{0:k}^{video}$, we use the ``modified description'' available in CaptainCook4D, which combines an original step description and its error description of how a user deviates from the corresponding recipe step. 

In the sampling, we sample around 200 \textit{next}, 200 \textit{missing}, and 20 other-type examples, approximately reflecting the total number of each type.

All the videos used in \dataset{} are from CaptainCook4D, which are released under Apache license 2.0.

\subsection{QA Generation}
\label{appx:qa-generation}

\autoref{fig:prompt-full-qa-generation} shows a full prompt example, and~\autoref{fig:prompt-dot} shows an example DOT representation of a recipe.

In the prompt exploration, the following are our findings: 
1) Feeding the recipe as a whole hurts the approval rate compared to the target excerpt. 
This can be because the LLM needs to do extra reasoning to identify where to focus on a recipe.
2) Feeding the video as frames worsens the approval rate. 
Videos contain more information to contextualize the generation.
However, the result suggests that even for the strong proprietary multimodal models, feeding information as text, if available, leads to better performance.
In addition, feeding as frames costs is much more expensive than feeding as text, as models require more tokens to process images. 

In the QA generator selection, we noticed that Gemini 1.5 Pro sometimes deviated from the specified format, \eg additional quotations or tags like ``\textit{[question]}'' or ``\textit{[answer]}''. 

\subsection{Verification}
\label{appx:verification}


\begin{table*}[!t]
\fontsize{8}{9}\selectfont
\setlength{\tabcolsep}{5pt}
\centering
\begin{tabular}{cccccccc|c}
    \toprule
    Missing & Next & Order & Measurement & Preparation & Technique & Temperature & Timing & Avg. \\
    \midrule    
    0.82 & 0.88 & 0.77 & 0.81 & 0.58 & 0.43 & 0.62 & 0.76 & 0.80 \\
    (148/180) & (158/179) & (20/26) & (17/21) & (14/24) & (9/21) & (10/16) & (25/33) & (401/500) \\
    \bottomrule
\end{tabular}
\caption{Approval rate (\#example after/before verification) for each question type.}
\label{tab:approval-rate}
\end{table*}

\begin{table*}[!t]
\centering
\begin{adjustbox}{width=0.9\textwidth}
\begin{tabular}{cccc}
    \toprule
    \makecell[c]{Annotator 1\\$\langle q^m, a^m_q, a^m_2, a^h_1 \rangle$} & \makecell[c]{Annotator 2\\$\langle q^m, a^m_q, a^m_2, a^h_2 \rangle$} & \makecell[c]{Adjudicator\\$\langle q^m, a^m_q, a^m_2, a^h_1, a^h_2, a^h_3 \rangle$} & Explanation\\
    \midrule
    $\langle$ \cmark{}, \cmark{}, \xmark{}, $\varnothing$ $\rangle$ & $\langle$ \cmark{}, \cmark{}, \xmark{}, $\varnothing$ $\rangle$ & $\langle$ --, --, --, --, --, -- $\rangle$ & \makecell[c]{Majority vote \& No Adjudication} \\
    \cmidrule(lr){1-4}
    $\langle$ \xmark{}, --, --, -- $\rangle$ & $\langle$ \xmark{}, -, -, - $\rangle$ & $\langle$ --, --, --, --, --, -- $\rangle$ & \makecell[c]{Majority vote \& No Adjudication}\\
    \cmidrule(lr){1-4}
    $\langle$ \cmark{}, \cmark{}, \cmark{}, $\varnothing$ $\rangle$ & $\langle$ \cmark{}, \cmark{}, \xmark{}, $\varnothing$ $\rangle$ & $\langle$ --, --, \cmark{}/\xmark{}, --, --, -- $\rangle$ & Majority vote\\
    \cmidrule(lr){1-4}
    $\langle$ \cmark{}, \cmark{}, \xmark{}, $\exists$ $\rangle$ & $\langle$ \cmark{}, \xmark{}, \xmark{}, $\exists$ $\rangle$ & $\langle$ --, \cmark{}/\xmark{}, --, \cmark{}/\xmark{}, \cmark{}/\xmark{}, -- $\rangle$ & \makecell[c]{Majority vote for $q^m, A^m$\\Adjudicator's call for $A^h$} \\
    \cmidrule(lr){1-4}
    $\langle$ \cmark{}, \cmark{}, \xmark{}, $\varnothing$ $\rangle$ & $\langle$ \xmark{}, --, --, -- $\rangle$ & \makecell[c]{$\langle$ \cmark{}, \cmark{}/\xmark{}, \cmark{}/\xmark{}, --, --, $\exists$ $\rangle$\\$\langle$ \xmark, --,--,--,--,--$\rangle$} & \makecell[c]{Majority vote for $q^m$\\Adjudicator's call for $A^m$\\Adjudicator can add $A^h$}\\
    \bottomrule
\end{tabular}
\end{adjustbox}
\caption{Case study of adjudicator's role. Suppose a QA generator generates a question $q^m$ and two answers $a_1^m, a_2^m$, and then, annotators optionally write human-written answers, $a_1^h$ by one annotator, $a_2^h$ by the other annotator, and $a_3^h$ by the adjudicator. The adjudicator's role changes based on two annotators' judges. (\cmark{}: valid, \xmark: invalid, $\varnothing$: no human-written answer, $\exists$: human-written answers exist, --: no judge added)}
\label{tab:adjudication-scenarios}
\end{table*}
\begin{table*}[!t]
\fontsize{8}{8}\selectfont
\setlength{\tabcolsep}{3pt}
\centering
\begin{tabular}{l c c c c c c c c c c c}
    \toprule
    \multirow{2}{*}{Model} & \multirow{2}{*}{Avg.} & \multicolumn{2}{c}{w/ Error} & \multicolumn{8}{c}{Question Type} \\
    \cmidrule(lr){3-4} \cmidrule(lr){5-12}
    & & clean & noisy & missing & next & order & measurement & preparation & technique & temperature & timing\\
    \midrule
    Llama 3.1 8B Instruct & 25.7 & 25.9 & 25.6 & 35.1 & 22.6 & 25.0 & 5.9 & 0.0 & 20.0 & 16.0 & 14.3 \\
    Llama 3.1 70B Instruct & 31.5 & 33.2 & 30.2 & 35.5 & 38.3 & 12.5 & 2.9 & 14.3 & 5.6 & 30.0 & 20.0 \\
    \midrule
    \makecell[l]{LLaVA 1.5 7B (50f, 288p)\\ \& Llama 3.1 8B Instruct} & 32.9 & 36.6 & 30.0 & 43.0 & 39.2 & 10.0 & 2.9 & 0.0 & 15.0 & 4.0 & 7.1 \\
     \makecell[l]{LLaVA 1.5 13B (50f, 288p) \\ \& Llama 3.1 70B Instruct} & 37.5 & 38.9 & 36.4 & 41.9 & 45.3 & 12.5 & 11.8 & 14.3 & 11.1 & 50.0 & 18.0 \\
    \midrule
    VideoLLaMA2 7B (8f, 336p) & 39.3 & 45.7 & 34.2 & 47.5 & 47.3 & 22.5 & 0.0 & 0.0 & 40.0 & 8.0 & 14.3 \\
    VideoLLaMA2 7B (16f, 336p) & 38.3 & 44.3 & 33.6 & 48.1 & 44.9 & 30.0 & 0.0 & 0.0 & 30.0 & 2.0 & 10.7 \\
    VideoLLaMA2 72B (8f, 336p) & 39.8 & 49.4 & 32.2 & 46.3 & 49.7 & 20.0 & 0.0 & 21.4 & 11.1 & 25.0 & 8.0 \\
    Qwen2 VL 7B (100f, 336p) & 33.8 & 38.6 & 30.0 & 43.4 & 36.8 & 30.0 & 0.0 & 0.0 & 30.0 & 6.0 & 14.3 \\
    Qwen2-VL 72B (100f, 336p) & 31.2 & 32.1 & 30.4 & 34.1 & 37.0 & 45.0 & 0.0 & 10.7 & 0.0 & 20.0 & 14.0 \\
    \midrule
    GPT-4o (50f, 765p) & 40.4 & 39.5 & 41.1 & 39.9 & 45.9 & 45.0 & 29.4 & 17.9 & 27.8 & 55.0 & 24.0 \\
    GPT-4o (100f, 288p) & 38.9 & 39.2 & 38.7 & 37.2 & 44.0 & 37.5 & 32.4 & 21.4 & 22.2 & 50.0 & 34.0 \\
    GPT-4o (250f, 288p) & 36.5 & 38.1 & 35.3 & 43.7 & 34.8 & 40.0 & 20.6 & 22.2 & 45.0 & 26.0 & 10.7 \\
    Gemini 1.5 Pro (50f, 765p) & 25.2 & 27.0 & 23.8 & 27.4 & 29.7 & 15.0 & 17.6 & 7.1 & 16.7 & 20.0 & 12.0 \\
    Gemini 1.5 Pro (100f, 288p) & 27.9 & 28.1 & 27.8 & 32.4 & 32.0 & 30.0 & 2.9 & 17.9 & 16.7 & 15.0 & 6.0 \\
    Gemini 1.5 Pro (250f, 288p) & 27.7 & 30.4 & 25.6 & 30.1 & 31.8 & 32.5 & 8.8 & 22.2 & 10.0 & 8.0 & 25.0 \\
    Claude 3.5 Sonnet (10f, 765p) & 44.1 & 48.9 & 40.4 & 44.6 & 58.2 & 27.5 & 8.8 & 14.3 & 5.6 & 25.0 & 28.0 \\
    Claude 3.5 Sonnet (100f, 288p) & 36.8 & 43.8 & 31.3 & 48.4 & 37.5 & 22.5 & 14.7 & 16.7 & 25.0 & 12.0 & 10.7 \\
    \midrule
    Human & 74.5 & 83.5 & 65.8 & --- & --- & --- & --- & --- & --- & --- & --- \\
    \bottomrule
\end{tabular}
\caption{Additional benchmark result: The average of all the examples, the averages of examples with (noisy) and without errors (clean) in previous steps, and the averages for the same question-type examples. 
f and p denote the number of frames and image resolution used for each model.
}
\label{tab:benchmark-additional-result}
\end{table*}
\begin{table*}[!t]
\fontsize{8}{8}\selectfont
\setlength{\tabcolsep}{3pt}
\centering
\begin{tabular}{l c c c c c c c c c c c}
    \toprule
    \multirow{3}{*}{Model} & \multirow{3}{*}{Avg.} & \multicolumn{3}{c}{Answer Source } & \multicolumn{5}{c}{Answer Type} \\
    \cmidrule(lr){3-5} \cmidrule(lr){6-10}
    & & machine & human & both & direct & direct \& suggestion & direct \& intervention & suggestion & all\\
    \midrule
    Llama 3.1 8B Instruct & 25.7 & 23.4 & 22.0 & 33.3 & 26.4 & 30.8 & 9.1 & 6.2 & 100.0 \\ 
    Llama 3.1 70B Instruct & 31.5 & 33.4 & 28.0 & 30.5 & 31.7 & 30.8 & 34.1 & 25.0 & 0.0 \\ 
    \midrule
    \makecell[l]{LLaVA 1.5 7B (50f, 288p)\\ \& Llama 3.1 8B Instruct} & 32.9 & 34.1 & 29.9 & 32.9 & 33.2 & 37.2 & 27.3 & 6.2 & 100.0 \\ 
     \makecell[l]{LLaVA 1.5 13B (50f, 288p) \\ \& Llama 3.1 70B Instruct} & 37.5 & 40.7 & 32.9 & 34.8 & 38.8 & 34.6 & 34.1 & 12.5 & 0.0 \\ 
    \midrule
    VideoLLaMA2 7B (8f, 336p) & 39.3 & 43.7 & 31.7 & 36.2 & 40.9 & 39.7 & 25.0 & 12.5 & 0.0 \\ 
    VideoLLaMA2 7B (16f, 336p) & 38.3 & 41.8 & 32.3 & 35.7 & 39.4 & 41.0 & 27.3 & 12.5 & 0.0 \\ 
    VideoLLaMA2 72B (8f, 336p) & 39.8 & 45.1 & 32.9 & 34.3 & 41.7 & 34.6 & 31.8 & 12.5 & 0.0 \\ 
    Qwen2 VL 7B (100f, 336p) & 33.8 & 36.7 & 28.0 & 32.4 & 35.3 & 28.2 & 29.5 & 12.5 & 0.0 \\ 
    Qwen2-VL 72B (100f, 336p) & 31.2 & 35.0 & 22.0 & 30.5 & 30.8 & 33.3 & 27.3 & 37.5 & 100.0 \\ 
    \midrule
    GPT-4o (50f, 765p) & 40.4 & 41.4 & 29.3 & 47.1 & 40.2 & 39.7 & 52.3 & 18.7 & 50.0 \\ 
    GPT-4o (100f, 288p) & 38.9 & 38.8 & 27.4 & 48.1 & 38.1 & 38.5 & 56.8 & 31.2 & 0.0 \\ 
    GPT-4o (250f, 288p) & 36.5 & 36.7 & 26.8 & 43.8 & 36.0 & 39.7 & 45.5 & 25.0 & 0.0 \\ 
    Gemini 1.5 Pro (50f, 765p) & 25.2 & 27.1 & 18.3 & 26.7 & 24.8 & 25.6 & 34.1 & 12.5 & 50.0 \\ 
    Gemini 1.5 Pro (100f, 288p) & 27.9 & 28.3 & 20.7 & 32.9 & 27.6 & 34.6 & 22.7 & 25.0 & 0.0 \\ 
    Gemini 1.5 Pro (250f, 288p) & 27.7 & 26.6 & 28.0 & 29.5 & 27.9 & 29.5 & 20.5 & 31.2 & 0.0 \\ 
    Claude 3.5 Sonnet (10f, 765p) & 44.1 & 48.1 & 35.4 & 42.9 & 45.0 & 42.3 & 43.2 & 25.0 & 0.0 \\ 
    Claude 3.5 Sonnet (100f, 288p) & 36.8 & 38.8 & 30.5 & 37.6 & 36.6 & 42.3 & 38.6 & 18.7 & 0.0 \\ 
    \bottomrule
\end{tabular}
\caption{
Additional answer-focused benchmark result breakdown: The average of all the examples, the averages of examples with only machine-generated answer(s), human-written answer(s), and both; The averages of examples with only direct answer(s), direct and suggestion(s), direct and intervention(s), only suggestion(s), and all answer-types. 
f and p denote the number of frames and image resolution used for each model.
}
\label{tab:benchmark-additional-result-answer}
\end{table*}

\autoref{fig:interface} shows the interface for verification. 
It consists of four parts: 
1) A recipe graph with step status (passed as green, current step as orange, and not passed as dotted) with the triangle on the upper left corner of each step indicates that it contains errors.
2) A recording.
3) A list of step and error descriptions. 
And, 4) QA annotation checkbox, including a comment box for human-written answers.
When a question is judged as valid, its answer checkboxes and comment box appear.

We distributed 500 examples to 5 annotators so that each example receives two annotators' judgments ($500 \times 2 = 1000$ judgments) and each pair of annotators (${}_{5}\mathrm{C}_{2} = 10$) shares 50 examples.
Based on the shared examples, we calculate the average of per-pair judgment agreements for both questions and answers, $0.87$, as discussed in~\autoref{ssec:verify}.
\autoref{tab:approval-rate} shows the breakdown approval rate for each question type. 
In general, GPT-4o generates more valid process-level questions than step-specific questions. 
Based on our manual inspection, one reason is that some error types are not suitable for multimodal questions. 
As shown in the table, \textit{preparation} and \textit{technique} produce less valid questions than others.
For instance, a step with an error description like ``The user peeled the onion improperly'' is unlikely to receive a multimodal question, as it is unlikely that recipes specify the detailed instructions.
Another potential reason is the quality of error descriptions in CaptainCook4D. 
Most of them are sensical, yet, they are not always grammatically correct, \eg dropping subjects, or detailed enough. 
Although we corrected the descriptions during our preliminary experiments, they were not exhaustive.

Also, we note that, in the training session for the verification, we received consent from the annotators about the potential release of their annotations.

\paragraph{Adjudication scenarios}
We set two base principles in designing the adjudicator's role: 
1) The adjudicator makes the final judgment for questions/answers when the judgments of two annotators conflict.
2) Every example receives two chances to receive human-written answers.
\autoref{tab:adjudication-scenarios} shows the role of the adjudicator in different cases. 
In the first three cases, all judgments are determined by a majority vote.
For the fourth one, while a machine-generated question and answers are judged based on a majority vote, human-written answers are judged and determined by the adjudicator's call. 
These human-written answers are the reasons why we have the two-stage verification process, \ie to have extra checks even for human-written answers.
Also, as two annotators can independently add human-written answers, there may exist duplicate answers, and we did not remove duplicates in the adjudication process.
Finally, only in the fifth case, the adjudicator can add human-written answers to comply with our second policy of two chances to receive human-written answers. 
We note that, as you can see from the case studies, not all examples have both machine-generated and human-written answers.
In the adjudication, we shuffle the order of answers in each example to make their sources (machine or human) unclear as a source could give extra bias to the adjudicator.

\section{Benchmarking}
\label{appx:benchmark}

\subsection{LLM-as-a-Judge}
\label{appx:llm-as-a-judge}

\autoref{fig:prompt-full-evaluation} shows one full prompt example for our LLM-based scoring.

\subsection{Experimental Details}
\label{appx:exp_setup}

We use \texttt{vllm} for the inference of Llama 3.1 and LLaVA 1.5. 
For VideoLLaMA2\footnote{\url{https://github.com/DAMO-NLP-SG/VideoLLaMA2}} and Qwen2-VL.\footnote{\url{https://github.com/QwenLM/Qwen2-VL}}, we follow their instructions to run their respective inference code.
All the weights are downloaded from \textit{HuggingFace}\footnote{\url{https://huggingface.co}} using \texttt{transformers}.
We use 1\textasciitilde{}4 GPUs of A6000 (48GB), depending on the size of the models.
Each inference took at most a few hours.
For all proprietary models, we use their libraries to make API calls. 
Each prediction on all of our 401 examples in \dataset{} costs 30\textasciitilde{}60 USD, depending on the model, the number of frames, and the resolution of each frame.
All the results are based on a single run.

\subsection{Additional Results}
\label{appx:benchmarking-additional-results}

\autoref{tab:benchmark-additional-result} and~\ref{tab:benchmark-additional-result-answer} show the additional benchmarking results by changing model size for open models and changing the number of frames and resolutions for proprietary models. 
The unimodal and Socratic models improve their performance as the sizes of their models increase. 
However, open multimodal models did not change the overall performance or even lowered their performance by a few points.
As for proprietary models, under the fixed maximum input length, the number of frames trades off the resolutions. 
In our experiment, we found that higher resolution leads to better performance. 
However, the combinations we tried are rather limited, and there may exist a better combination, which we leave the exploration for future work.
We also note here that we faced issues with limited maximum lengths with image-included prompts, compared to the ones listed on each API documentation or the ones when we tried text-only prompts. 
Presumably, this is due to the large file size of each image and the total data size of one input for each API request. 
We also leave it for future work on the workaround of how to feed many relatively high-resolution images in input prompts. 

\begin{figure*}
    \begin{lstlisting}
# Instruction
A person is cooking Spiced Hot Chocolate with their friend, who is a skilled cook.  
The person completed these steps:
- Fill a microwave-safe mug with whole milk but spill
- Microwave the contents of the mug for 2 minutes
- Add-Add 4 pieces of chocolate to the mug
- Add-Add 1 teaspoon of white sugar to the mug
And, the person has just performed this step:
- Mix-Mix the contents of the mug
The friend knows the following step(s) can be done next:
- Heat-Heat the contents of the mug for 1 minute and serve 
The person may or may not be noticing this.
What questions would the person ask the friend about next step(s)? 

Assuming the friend is watching over you throughout the cooking activity and 
understand the situation, return three pairs of a question and its answers as a 
list:
* <questions>
    * <answer1-1>                                                 
    * <answer1-2>                                                 
    * ...                                                       
* <question2>                                                     
    * ...

# Note
- Each question/answer should consists of one consice sentence/phrase.
- If there exist multiple correct answers, provide all correct answers for each question as a list so that each answer targets at one step.
- Each answer targets at one step.                                - Imagine a variety of a person: beginner/experienced, careless/careful, etc...
- It is preferable to have as diverse pairs (question/answer type, tone, wording, etc) as possible.
- There is a case where no missing step is performed, i.e., an answer is just no.

# Example
* What is the next step?
    * You have completed all the steps.
* What should I do next?
    * <stepY>
    * <stepZ>
    
# Response
    \end{lstlisting}
    \caption{Prompt example for QA generation: $next$ question}
    \label{fig:prompt-full-qa-generation}
\end{figure*}
\begin{figure*}
    \begin{lstlisting}
digraph G { 
    START; "Heat-Heat the contents of the mug for 1 minute and serve"; 
    "Add-Add 1/5 teaspoon cinnamon to the mug"; 
    "Mix-Mix the contents of the mug"; 
    "Add-Add 1 teaspoon of white sugar to the mug"; 
    "Fill-Fill a microwave-safe mug with skimmed milk"; 
    "Microwave-Microwave the contents of the mug for 1 minute"; 
    "Add-Add 2 pieces of chocolate to the mug"; 
    END; 
    
    "Mix-Mix the contents of the mug" -> "Heat-Heat the contents of the mug for 1 minute and serve"; 
    "Add-Add 2 pieces of chocolate to the mug" -> "Mix-Mix the contents of the mug"; 
    "Add-Add 1 teaspoon of white sugar to the mug" -> "Mix-Mix the contents of the mug"; 
    "Add-Add 1/5 teaspoon cinnamon to the mug" -> "Mix-Mix the contents of the mug"; 
    "Microwave-Microwave the contents of the mug for 1 minute" -> "Add-Add 1 teaspoon of white sugar to the mug"; 
    START -> "Fill-Fill a microwave-safe mug with skimmed milk"; 
    "Heat-Heat the contents of the mug for 1 minute and serve" -> END; 
    "Microwave-Microwave the contents of the mug for 1 minute" -> "Add-Add 2 pieces of chocolate to the mug"; 
    "Microwave-Microwave the contents of the mug for 1 minute" -> "Add-Add 1/5 teaspoon cinnamon to the mug"; 
    "Fill-Fill a microwave-safe mug with skimmed milk" -> "Microwave-Microwave the contents of the mug for 1 minute"; 
}
    \end{lstlisting}
    \caption{Prompt example of a recipe in DOT format: ``Spiced Hot Chocolate''}
    \label{fig:prompt-dot}
\end{figure*}
\begin{figure*}
    \begin{lstlisting}
# Instruction
This is an evaluation task.
You will be given a question, gold answer(s), and predicted answer.
Your task is to evaluate if the predicted answer matches against the gold answer(s).

Give your ternary judge 0, 1, or 2:
* 0 means the predicted answer is wrong (unmatch)
* 1 means the predicted answer is partially correct/wrong (partial match)
* 2 means the predicted answer is correct (match)
When multiple gold answers are available (provided as a list), the predicted answer is correct/partially correct if it matches/partially matches with at least one of the gold answers.

Provide your feedback as follows:
# Feedback
[Rationale] (your rationale for the judge, as a text)
[Judge] (your judge, as a number, 0, 1, or 2)

# Note
The question is being asked by a user who is cooking Cucumber Raita.
Well-trained annotators constructed gold answer(s), while the predicted answer was by a machine, which answered based on the corresponding recipe and the frames of the cooking recording.

Here are the steps being performed already:
- Add-Add 1 teaspoon of cumin powder to the bowl
- add-add 1 tablespoon of chopped scallions to the bowl instead of cilantro
- Rinse-Rinse 1 medium sized zucchini
- Add-1/4 teaspoon of red chilli powder to the bowl
- whisk-In a mixing bowl, whisk 1 cup of chilled curd until smooth. Use fresh homemade or packaged curd
- chop or grate-chop or grate only 1/2 of zucchini instead of one medium cucumber

# Task
Now, here are the question, gold answer(s), and predicted answer:
[Question]
- Did I forget any other ingredients?
[Gold Answer(s)]
- No, you did not forget any ingredients at the moment.
[Predicted Answer]
- Based on the images, it seems you forgot to add 1/2 teaspoon of chaat masala powder.

# Feedback
[Rationale]
    \end{lstlisting}
    \caption{Prompt example for evaluation.}
    \label{fig:prompt-full-evaluation}
\end{figure*}

\end{document}